\pdfoutput=1

\documentclass[11pt]{article}

\usepackage[]{acl}

\usepackage{times}
\usepackage{latexsym}

\usepackage[T1]{fontenc}

\usepackage[utf8]{inputenc}

\usepackage{microtype}

\usepackage{booktabs}
\usepackage{graphicx}
\usepackage{multirow}
\usepackage{adjustbox}  
\usepackage[symbol]{footmisc}

\title{Framing in the Presence of Supporting Data: A Case Study in U.S. Economic News}

\author{
   Alexandria Leto$^{1}$ \,\,
   Elliot Pickens$^{2}$   \,\,
   Coen D. Needell$^{3}$ \\
   \textbf{David Rothschild}$^4$   \,\,
   \textbf{Maria Leonor Pacheco}$^{1}$ \\
   $^1$University of Colorado Boulder\,\,
   $^2$University of Wisconsin Madison\\
   $^3$University of Pennsylvania \,\,
   $^4$Microsoft Research \\
    $^1$\texttt{\{alexandria.leto, maria.pacheco\}@colorado.edu} \\
   $^2$\texttt{epickens@cs.wisc.edu} \,\, $^3$\texttt{coen@needell.org} \,\, $^4$\texttt{david@researchdmr.com}
}

\begin{document}
\maketitle
\begin{abstract}
The mainstream media has much leeway in what it chooses to cover and how it covers it. These choices have real-world consequences on what people know and their subsequent behaviors. However, the lack of objective measures to evaluate editorial choices makes research in this area particularly difficult. In this paper, we argue that there are newsworthy topics where objective measures exist in the form of supporting data and propose a computational framework to analyze editorial choices in this setup. We focus on the economy because the reporting of economic indicators presents us with a relatively easy way to determine both the selection and framing of various publications. Their values provide a ground truth of how the economy is doing relative to how the publications choose to cover it. To do this, we define frame prediction as a set of interdependent tasks. At the article level, we learn to identify the reported stance towards the general state of the economy. Then, for every numerical quantity reported in the article, we learn to identify whether it corresponds to an economic indicator and whether it is being reported in a positive or negative way. To perform our analysis, we track six American publishers and each article that appeared in the top 10 slots of their landing page between 2015 and 2023.

\end{abstract}

\section{Introduction}


The mainstream media has much leeway in what it chooses to cover and how it covers it. Should the top article at any given point in time be about an international incident or the state of the economy? Conditional on choosing to cover the economy, should the article focus on the indicator that is up or the one that is down? Should it be optimistic or pessimistic towards the future? These choices have real-world consequences affecting outcomes such as voting behavior~\cite{0c0cf6378cd0467387ccfef508f5af0f,10.1257/app.1.2.35}, gun purchases~\cite{krupenkin2023fear}, and attitudes towards immigrants~\cite{krupenkin2020beyond, doi:10.1177/19401612231204535}. Most selection and framing decisions lack objective measures, 
making research on this topic difficult. 

\begin{figure}
    \centering
    \includegraphics[width=\columnwidth]{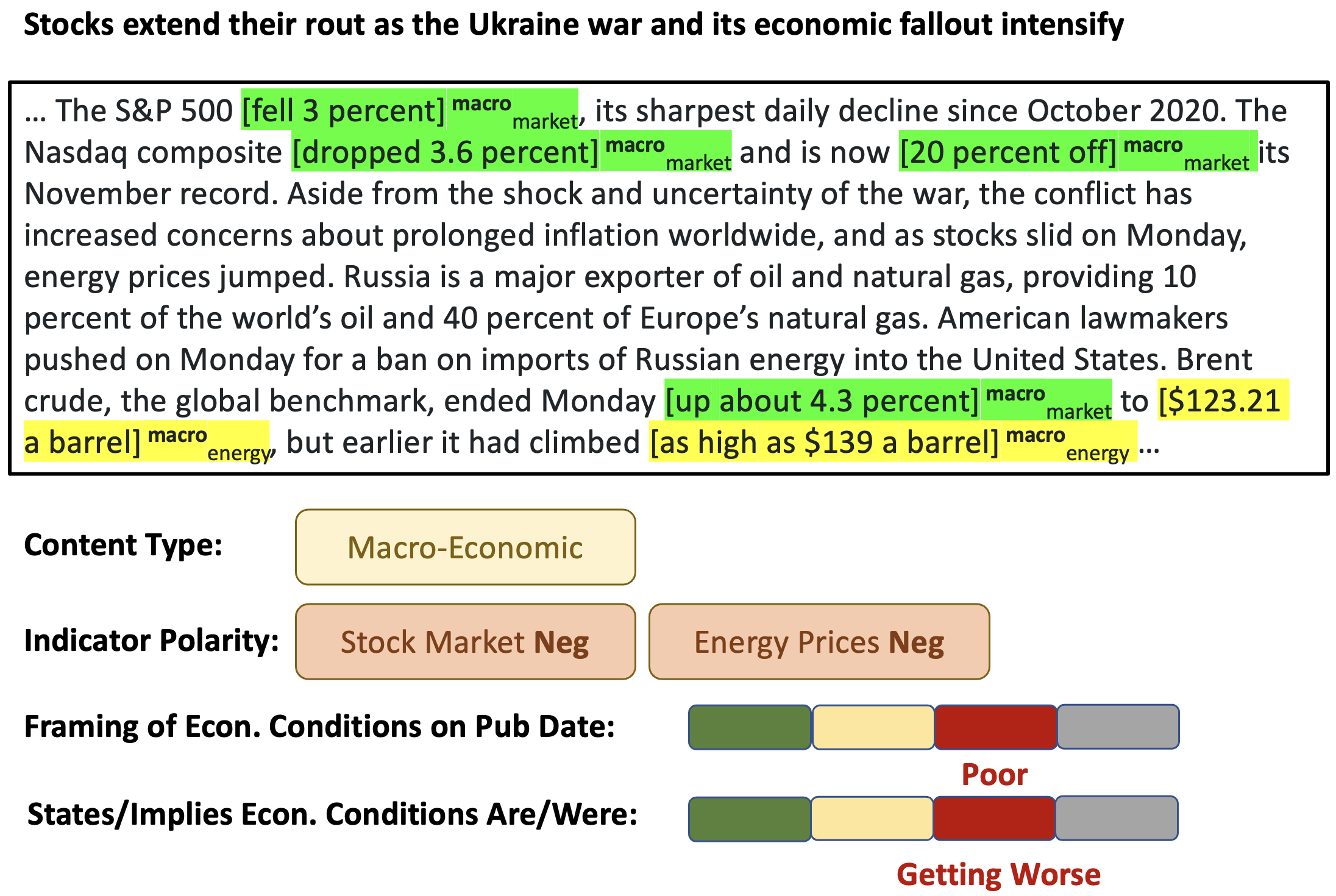}
    \caption{The Frame Prediction Framework}
    \label{fig:task}
\end{figure}

In this paper, we make the observation that there are a select number of newsworthy topics where objective measures \textit{do} exist. For example, there is a finite set of economic indicators that experts use to define the state of the economy, such as changes in non-farm payroll and gross domestic product. Another example is crime, which has well-established data points maintained by the FBI and local jurisdictions. 
Building a system capable of tracking a large number of publications, identifying \textit{which} indicators are reported and \textit{how} they are being reported has enormous research potential. Such a system would enable us to create three views of potential bias: how a publication is covering a topic over time, how their coverage differs from other publications, and how it compares to the accepted ground truth of experts in the field.

Following this rationale, we propose a computational framework to predict frames \textit{in the presence of supporting data}. For a given article, our goal is to automatically identify \textit{how} the general topic is being portrayed, \textit{which} indicators are being reported to support this view, and \textit{how} each of these indicators are being presented. We focus specifically on news articles about the U.S. economy. However, we argue that this framework could be adapted to other domains where numerical indicators are used, including reports on crime, climate change, and public opinion.

Most previous computational framing analysis approaches have conceptualized and interpreted frames as high-level relevant topics and themes \cite{ali-hassan-2022-survey}. For example, \citet{Boydstun2014} proposed 15 broad dimensions that aim to capture ways in which policy issues are discussed in news articles. The dimensions include themes such as ``economic'' and ``public opinion''. These broad dimensions fail to capture a frame's nuances. For example, the ``economic frame'' focuses on anything related to the economy, which is insufficient to answer the question of \textit{how} different aspects of the economy are being presented. 

To address this problem, we decompose economic frames into a set of interdependent tasks (See Fig. \ref{fig:task}). At the article level, we identify the relevance of the article to the economy, the type of economic information it covers (e.g., macro economic, industry-specific, firm-specific), and two measures of general economic sentiment: the state of the economy (e.g., good, fair, poor), and the direction in which the economy is heading (e.g., better, same, worse). Then, for each numerical quantity reported in the article, we predict whether it corresponds to an economic indicator (e.g., market numbers) and its reported polarity (e.g., “The S\&P 500 fell 3\%” has a negative polarity). This decomposition captures three levels: whether the economy is being discussed, which economic indicators are highlighted, and the framing of those indicators and the economy as a whole. This way, we can go beyond analyzing topic selection by additionally measuring how a publication's reporting differs from the objective reality at that time, how it evolves across time, and how it compares to other publications. We present a computational model that predicts each proposed frame component jointly. It utilizes the inter-dependence between the sub-tasks (e.g., polarity of quantities and polarity of the article they appear in) to improve upon results obtained by independent classifiers trained to make predictions individually with limited supervision.





We make the following contributions: (1) We propose a computational framework to model framing in the presence of supporting data and apply it to news about the U.S. economy. (2) We collect a novel dataset of landing page news articles published by a set of major U.S. news outlets between 2015 and 2023, and provide high-quality annotations of our proposed frame components for a small subset of examples. (3) We propose an automated method to detect each of the proposed frame components under low-supervision settings. (4) We demonstrate that our framework can be used to track \textit{when} and \textit{how often} (selection), as well as \textit{how} (framing) different news outlets report on specific aspects of the economy. We compare these reports to the ground truth for two example indicators: job numbers and prices. All of our code and data has been released to the community\footnote{\url{https://github.com/blast-cu/econ-indicators}}.
\section{Background and Related Work}



In this section, we cover the background on selection and framing, explain economic indicators, and discuss previous approaches to analyze framing. 

\paragraph{Selection and Framing} National news publications have a massive choice-set of topics they could cover on any given day. Though it may seem like the publications are reacting to a defined set of events that recently happened, outside of a few major events, they choose what the news is on any given day. In other words, the news in mainstream publications does not present an objective view of current affairs. Rather, it is influenced by the \textit{selection} made by media professionals~\cite{mcquail1992media}. Selection is driven by what researchers have called news values \cite{doi:10.1177/002234336500200104,values_revisited,values_revisited_again}, which determine whether a story is newsworthy. 

While there is variation in the taxonomies that have been proposed to determine newsworthiness, the most recent ones include factors such as: the power of the elite (stories concerning powerful entities), bad and good news (stories with particular emotional overtones), and magnitude (stories that are perceived as significant for a wide audience) \cite{values_revisited_again}. 
More interestingly, news values vary from publication to publication. A study of the run-up to the 2022 election showed that the \emph{New York Times} and \emph{Washington Post} averaged just two overlapping topics each morning in their printed editions out of about six articles \cite{rothschild2023}.

Conditional on selecting a topic, there is additional leeway on how to cover it, as there is a wide-range of facts and opinions that could spin the takeaway of the readers. This phenomenon, referred to as \textit{framing}, has been widely studied in communication studies. While there is not an agreed upon definition of framing, recent surveys~\cite{vallejo2023connecting} have identified the following prevalent definitions: equivalence framing (presenting the same exact information in different ways)~\cite{cacciatore2015}, emphasis framing (highlighting specific aspects of an event to promote a particular interpretation)~\cite{entman2007}, and story framing (leveraging established narratives to convey information)~\cite{hallahan1999}. 

When it comes to the economy, a publication could present a rosy or terrible version of the economy. They can do so without lying by simply picking the right economic indicator with the right perspective. For example, there are months where the value of the non-farm payroll is above average, but below the market expectation: either perspective is technically true. But, with a relatively finite set of indicators, and an established way for economists to judge them, there is still a clear ground truth to compare if any given indicator with any given perspective represents the consensus understanding of the economy at that time.

\paragraph{Economic Indicators} Understanding the health and wealth of the U.S. economy is critical for investment and growth. Most modern economic indicators have been tracked for decades, codified in the by the mid-20th century. Examples include: jobs numbers (e.g., non-farm payroll), prices (e.g., consumer price index), and macro economic (e.g., gross domestic product). With the exception of market prices (e.g., S\&P500), which are a reflection of the underlying assets, the core indicators are tracked by a few government agencies (e.g, the Bureau of Economic Analysis for GDP, spending, income, trade, and the Bureau of Labor Statistics for unemployment, prices, productivity), which conduct regular, large scale polling.

These economic indicators inform the public on the health of the economy. The job numbers are front page news when they are released at 8:30 AM ET on the first Friday of each month. The state of the markets is constantly covered in the news. Indicators such as the consumer price index pop up whenever the media finds them most important, interesting, and engaging. Most people have limited views of the economy outside of their close social circle, so these values are the key to their understanding of the health of the economy as a whole. And, traditionally, the health of the economy has tracked \textit{very closely} with the approval of incumbent political leaders~\cite{hummel2014fundamental}.

\paragraph{Framing of Economic News} This paper builds off of an extensive literature that has not just identified the quantity of economic coverage, but framing of that coverage. Most of this work is geared towards matching the sentiment of articles about the economy to both consumer sentiment and economic indicators, tracking both historical value and predicting future ones \cite{hopkins2017does, ardia2019questioning, shapiro2022measuring, seki2022news, bybee2023ghost, NBERw32026}. While the methods continue to evolve from simple keywords to BERT or LLM-based predictions, they are focused on article-level sentiment, aggregated to states or the U.S. They also study a mix of questions about economic impact that benefit from a long time-series of sentiment, granular in both time and geography, regarding questions as diverse as the impact of bubbles on industries \citep{bybee2023ghost}, economic shocks to economic decisions \cite{shapiro2022measuring}, to the relationship between news and perception of the economy \cite{hopkins2017does}. Our paper extends this literature by capturing not only the sentiment of economic articles, but what economic indicators and economic indicator-level sentiment is driving it. This allows us to dissect what editorial choices are driving perceived sentiment around the economy.

\paragraph{Computational Framing Analysis}
Scholars have increasingly adopted computational approaches to study framing, allowing studies to scale to large media repositories.
In most cases, researchers adopt unsupervised techniques such as topic modeling to identify latent themes~\cite{dimaggio2013,nguyen-etal-2015-tea,RePEc:wly:amposc:v:65:y:2021:i:1:p:21-35}. However, topic models are limited in their ability to capture framing nuances. By defining topics as simple distributions over words, they lack the semantic and discursive conceptualization needed to answer \textit{how} issues and events are being presented~\cite{ali-hassan-2022-survey}.

There is also a body of work leveraging supervised learning~\cite{johnson-etal-2017-leveraging,khanehzar-etal-2019-modeling,10.1145/3394231.3397921,huguet-cabot-etal-2020-pragmatics,mendelsohn-etal-2021-modeling} and lexicon expansion techniques~\cite{field-etal-2018-framing,roy-goldwasser-2020-weakly} to analyze framing. For this to be feasible, authors must define a concrete taxonomy of which frames are relevant and how they are represented in data. Most of these approaches rely on the taxonomy proposed by ~\citet{Boydstun2014}. These dimensions correspond to broad themes such as ``economic'', and ``public opinion''. While these themes have shown to be useful to model a select set of contentious political issues like abortion and immigration, they are too broad to capture \textit{how} the economy and other data-driven topics are being presented in the media.

\section{Data}\label{sec:data}


In this section, we describe our data collection process, annotation schema, annotation guidelines, and quality assurance process. We also present statistics for our resulting dataset. 

\paragraph{Data Collection}



The data for this study came from the Internet Archive's Wayback Machine \cite{Wayback}. We consider six American publishers: The New York Times, The Wall Street Journal, The Washington Post, Fox News, HuffPost and Breitbart. Selected sources represent a broad range of mainstream media outlets that were easily accessible through the archive. Breitbart is included for a future analysis comparing mainstream media with a fringe source. 
%
For each publisher, we collected the ``front page'' from every Wayback Machine entry between Jan. 1st 2015 to Jan. 1st 2023.
For each front page, we 
recorded articles in the top 10 positions and discarded all duplicates. 

To identify articles that discuss the economy, we curated a lexicon of economic terms sourced from the Bureau of Labor Statistics and the FRED database operated by the Federal Reserve bank of St. Louis.\footnote{\url{https://www.bls.gov/data/}}\footnote{\url{https://fred.stlouisfed.org/}} These sources provide a fairly comprehensive list of common metrics for economic activity. 
We consider an article to be relevant to the economy if it contains at least three sentences mentioning an economic term from our lexicon. This resulted in a total of 199,066 articles (Tab.~\ref{tab:data_stats}). App. \ref{app:data_collection} includes additional details about the data collection process and the lexicon of economic terms. 

\paragraph{Data Annotation}\label{Data-annotation}

We employed a group of six annotators to label our dataset following the guidelines outlined in App. \ref{app:ann_guidelines}. Our annotators comprised senior, postdoctoral and pre-doctoral researchers. Their fields of study include: economics, computational social science, and communications. They received training from the authors of the paper and several rounds of calibration were performed before annotation began. 
Articles were chosen such that there would be some overlap between coders (in order to check quality) and a decent breadth of coverage. We prioritized topic diversity over uniformity of news outlets and thus followed the sampling procedure introduced in ~\citet{pacheco-etal-2022-interactively, pacheco-etal-2023-interactive}.

\begin{figure}[t]
    \centering
    \includegraphics[width=\columnwidth]{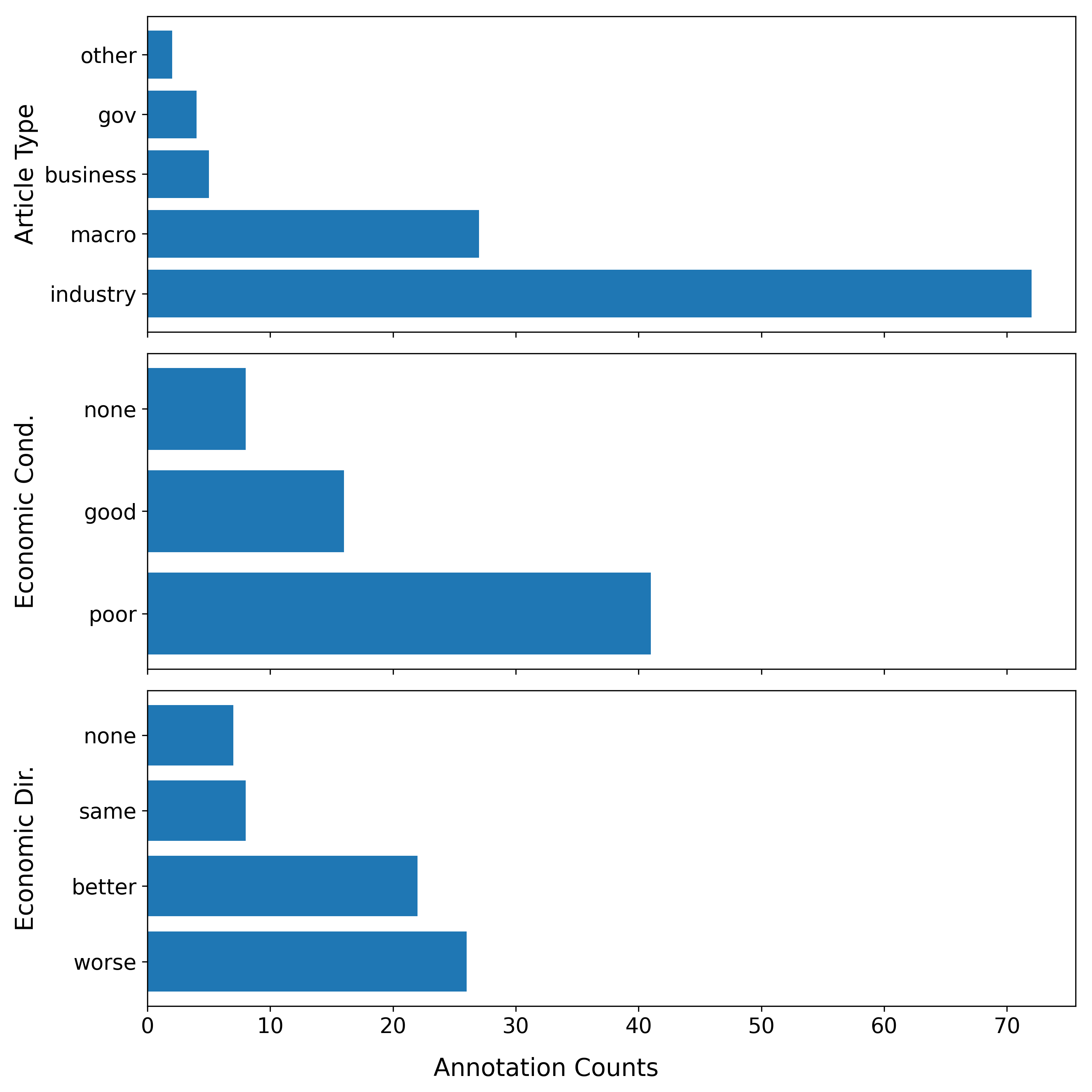}
    \caption{Distribution of article-level annotation labels}
    \label{fig:label_dist_qual}
\end{figure}
\begin{figure}[t]
    \centering
    \includegraphics[width=\columnwidth]{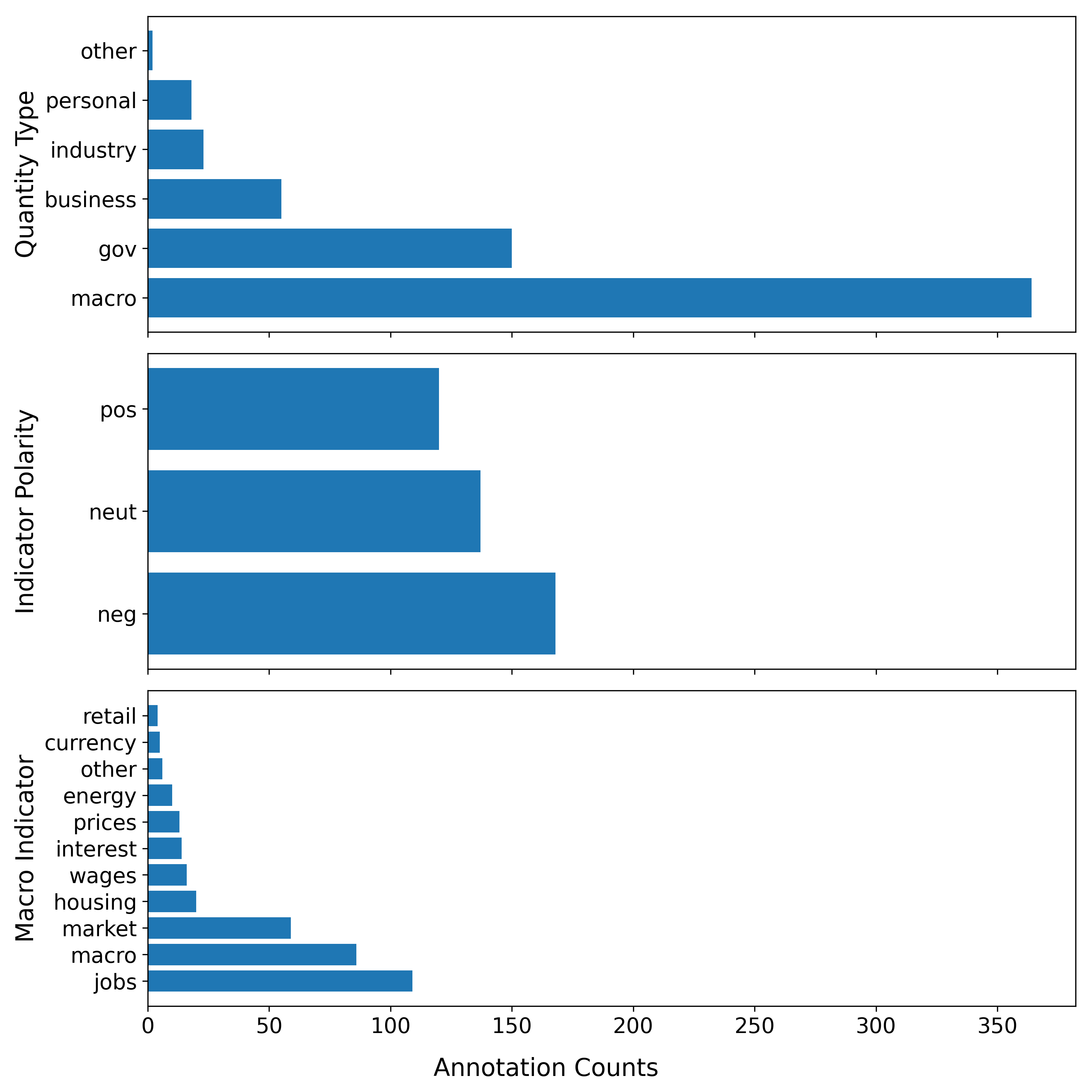}
    \caption{Distribution of quantity-level annotation labels}
    \label{fig:label_dist_quant}
\end{figure}

For each article in their batch, annotators selected: (1) what \textit{type} of economic information was the most prominent (macro, government, industry-specific, business-specific or personal), (2) the framing of the general economic \textit{conditions} (good, fair, bad), and (3) the framing of the \textit{direction} that the economy is heading (better, same, worse). These questions were adapted from the Gallup economic index~\footnote{\url{https://news.gallup.com/poll/1609/consumer-views-economy.aspx}}. Then, for every valid numerical value reported in the article, annotators selected: (1) what type of economic data it reports (macro, government, industry-specific, business-specific or personal), (2) which indicator was reported (e.g., jobs, prices. See App. \ref{app:ann_guidelines} for the full list of options), and (3) the reported polarity of each quantity (positive, negative, neutral). 
We report resulting statistics in Table~\ref{tab:data_stats}, and inter-annotator agreement in Tab. ~\ref{tab:ann_agreeement}. To calculate agreement, we only consider examples that were annotated by two or more coders. In App. \ref{app:cross_ann_stats} we include statistics for the number of coders for each of the label categories. Lastly, we show the resulting annotation distribution for all frame components in Figs.~\ref{fig:label_dist_qual} and ~\ref{fig:label_dist_quant}.

\begin{table}
    \centering
    \resizebox{\columnwidth}{!}{%
    \begin{tabular}{lrr|rr}
        \toprule
        \multirow{2}{*}{Publisher} & Econ  & \multirow{2}{*}{Quants.} & \multicolumn{2}{c}{Human Annotations}  \\
        \cline{4-5}
        & Arts. & & Art-level  & Quant-level \\
        \midrule
         New York Times & 58,240 & 370,723 & 516 & 1,117\\
         Wall Street Journal & 46,267 & 551,726 & 206 & 493\\
         Washington Post & 44,016 & 274,197 & 231 & 421\\
         Fox News & 21,795 & 76,074 & 45 & 42\\
         Breitbart & 15,954 & 66,572 & 91 & 149\\
         HuffPost & 12,794 & 75,180 & 82 & 210\\
         \midrule
         Total & 199,066 & 1,414,472 & 1,171 & 2,414    \\
         Cross-Annotated & & &  270 &   689\\
         
         \bottomrule
    \end{tabular}}
    \caption{Resulting Dataset Statistics}
    \label{tab:data_stats}
\end{table}

\begin{table}\small
    \centering
    \begin{tabular}{ll|ll}
    \toprule
    \multirow{2}{*}{Annotation} & \multirow{2}{*}{K's $\alpha$} & \multicolumn{2}{c}{\% of Agreement} \\
    & & Full & Partial \\
    \midrule
        Article Type &  0.48 & 57.81 & 90.62  \\
        Econ. Conditions & 0.43 & 55.13 & 85.90 \\
        Econ. Direction & 0.42 & 47.44 & 82.05 \\
        \midrule
        Quantity Type & 0.75 & 80.84 & 88.97 \\
        Quantity Polarity & 0.56 & 68.04 & 76.48 \\ 
        \;\;\;\; Macro Ind. & 0.83 & 83.42 & 92.78 \\
        
    \bottomrule
    \end{tabular}
    \caption{Inter-annotator Agreement for all examples with 2 or or more annotators. Here, Krippendorff's $\alpha = 1$ indicates perfect reliability, $\alpha = 0$ indicates the complete absence of reliability, i.e., judgments are random, and $\alpha < 0$ indicates that disagreements are systematic and exceed what can be expected by chance.}
    \label{tab:ann_agreeement}
\end{table}

We find that inter-annotator agreement is generally good for quantity-level information (Krippenforff's $\alpha$ 0.56-0.83). This reflects our intuition that capturing framing at the level of supporting data points may be a better alternative to relying on general topical markers. On the other hand, article-level judgements which aim to capture main ``takeaways'' are considerably harder for annotators, exhibiting moderate agreement (Krippenforff's $\alpha$ 0.42-0.48). This is expected and in line with other high-level framing tasks~\cite{card-etal-2015-media,roy-etal-2021-identifying,mendelsohn-etal-2021-modeling}. 
To further characterize the lower agreement values, we include confusion matrices and some examples illustrating ambiguous cases in App.~\ref{app:cm_ann}. We find that most of the confusion lies on the following article-level judgements: (1) identifying the main type of the article, particularly deciding between macro-economic and government-specific types when both types of data points are present in the article, (2) characterizing articles that are positive towards the economy, as they tend to be more subtle than the clearly negative ones, and (3) identifying articles that suggest that the economy is doing neither better nor worse, but staying the same, which is an inherently ambiguous category. We have included a similar confusion analysis for quantitative-level judgements in App.~\ref{app:cm_ann}.


\section{Model}

Obtaining high-quality annotations for our task is time-consuming and requires considerable domain expertise. For this reason, one of the main challenges that we face in predicting economic frames is a small amount of supervision. To circumvent this constraint we combine two modeling strategies: (1) Exploiting the decomposition of frame prediction into a set of modular, inter-dependent sub-tasks, and (2) Leveraging pre-training strategies using our large set of in-domain unlabeled examples (all 199,066 economic articles collected). 

\paragraph{Exploiting inter-dependencies with SRL} Statistical relational learning (SRL) methods attempt to model a joint distribution over relational, inter-dependent data~\cite{richardson06markov,10.5555/3122009.3176853,pacheco-goldwasser-2021-modeling}. These methods have proven particularly effective in tasks where contextualizing information and interdependent decisions can compensate for a low number of annotated examples \cite{deng-wiebe-2015-joint,roy-etal-2021-identifying,pacheco-etal-2022-holistic}. 

To model framing, we take advantage of the relations between the values of the frame components for a given article and implement a relational model using Probabilistic Soft Logic (PSL)~\cite{10.5555/3122009.3176853}. 
Dependencies in PSL are expressed using weighted logical rules of the form: $w_r : P_1 \wedge ... \wedge P_{n-1} \rightarrow P_n$, where $w_r$ indicates the importance of the rule in the model, and can be learned from data. Predicates $P_i$ correspond to decisions and observations. Rules are  then compiled into a Hinge-Loss Markov random field and weights are learned using maximum likelihood estimation. 
We consider the following rules:

\textbf{Priors.} We explicitly model prior probabilities for all article-level (type, economic conditions and economic direction) and quantity-level (type, macro-indicator, and sentiment) components. These priors are derived from supervised classifiers trained on our labeled dataset.

\textbf{$(r_1)$ Consistency between the quantity type and the macro-indicator.} We enforce that if the quantity of a type is macro, then a macro-indicator must be predicted, and vice-versa: $\mathtt{Reports(a,q)} \wedge \mathtt{Type(q,macro)} \Leftrightarrow \neg \mathtt{MacroIndicator(q,none)}$

\textbf{$(r_2)$ Consistency between the article type, economic conditions and direction.} We enforce that if the article type is macro, then economic conditions and economic direction must be predicted and vice-versa: $\mathtt{Type(a,macro)} \Leftrightarrow \neg \mathtt{EconConditions(q,irrelevant)} \wedge \neg \mathtt{EconRating(q,irrelevant)}$

\textbf{$(r_3)$ Dependency between the polarity of quantities and economic conditions.} Our intuition is that if several negative quantities are reported in the article, then the article is likely to frame economic conditions as negative, and vice-versa for positive quantities: $\mathtt{Reports(a,q)} \wedge \mathtt{Polarity(q,p)} \Rightarrow  \mathtt{EconConditions(a,p)}$

\textbf{$(r_4)$ Dependency between the polarity of quantities and economic direction.} Our intuition is that if many negative quantities are reported in the article, then the article is likely to frame the direction of the economy as negative, and vice-versa for positive quantities: $\mathtt{Reports(a,q)} \wedge \mathtt{Polarity(q,p)} \Rightarrow \mathtt{EconDirection(a,p)}$

\textbf{$(r_5)$ Dependency between neighboring quantity types.} Our intuition is that there are common patterns in the types of consecutive quantities (e.g., sequential dependencies): $\mathtt{Precedes(q1,q2)} \wedge \mathtt{Type(q1,t1)} \Rightarrow \mathtt{Type(q2,t2)}$


Rule $r_1$ and $r_2$ are modeled as \textit{hard} constraints, and are always enforced because they are designed to force predictions to follow the frame annotation structure. Rules $r_{3-5}$ are modeled as \textit{soft} constraints, and their weights are learned through PSL. This way, we allow for predictions that do not conform to the templates to be active.

\paragraph{Enhancing PSL priors with pre-trained language models} Pre-trained language models (LMs) are one of the most effective techniques for acquiring knowledge from unlabeled text data~\cite{devlin-etal-2019-bert}. Domain-adaptive pre-training (DAPT) further enriches LMs with in-domain data~\cite{gururangan-etal-2020-dont}. In both of these pre-training stages, learning is conducted by randomly masking words in a large dataset and training the the LM to predict these words. To take advantage of these strategies, we use pre-trained RoBERTA and perform DAPT with our large unlabeled dataset (all 199,066 economic articles collected). We use this LM to fine-tune the classifiers that are used as priors in PSL. This strategy has been repeatedly used to combine PSL with strong classifiers, with consistent success ~\cite{sridhar-etal-2015-joint, pacheco-goldwasser-2021-modeling, roy-etal-2021-identifying}.

\section{Experimental Evaluation}\label{sec:experiments}
In this section, we outline our experimental settings and present results for our  frame prediction model. We also include an example of the type of framing analysis that can be done using our model predictions over the complete dataset. 

\subsection{Frame Prediction}\label{sec:prediction}
To determine the effectiveness of exploiting inter-dependencies with statistical relational learning and our additional pre-training steps, we conducted an ablation study in addition to evaluating the model as a whole. We provide a discussion of the results of these experiments, in which we detail per-class and per-publisher metrics.

\paragraph{Experimental Settings} We perform 5-fold cross-validation in all scenarios, and assume a multi-class setup for each frame component. To create our folds, we first split the articles that were cross-annotated and where inter-annotator agreement was reached (270 articles, 689 quantities) into five folds, and consider 4 folds for training and 1 for testing. Then, we further enhance the training data for all cases with the additional annotated articles, which we consider as a source of noisy supervision. More details about the data splits used for experiments are included in App. \ref{app:setup}. 

The base classifiers are initialized with pre-trained RoBERTA (with and without DAPT) and trained using the AdamW optimizer, cross-entropy loss, and a learning rate of $2e-5$. For early stopping, we use the macro F1 on the dev set, consisting of ten percent of the training articles for each fold. For more details on the classifier architectures and training settings, see App. \ref{app:setup}. Predictions are fed into PSL as priors, and rule weight learning is done with the standard configuration. We report the average macro F1 scores for all folds.  

\paragraph{Results} We present our general results in Tab.~ \ref{tab:general_macro}. This includes baselines obtained by zero-shot and two-shot prompting \texttt{Mistral-7b-Instruct-v0.2}. Additional details about the prompts and performance can be found in App. \ref{app:setup}. We find that fine-tuning, even with relatively little data, results in better performance than prompting an LLM. Additionally, we find that DAPT is considerably helpful for quantity-level predictions, but does not improve article-level predictions. We hypothesize that it is difficult for RoBERTa to model the long context of articles to take advantage of the masked language modeling objective. Next, we find that the relational model gives us a clear boost in performance over the base classifiers, with the exception of article type. This improvement supports our modeling decision to decompose frame prediction into inter-dependent sub-tasks. While Tab.~\ref{tab:general_macro} reports macro F1, we find the same trend for weighted F1 (See App. \ref{app:additional}).


\begin{table}[t]
    \centering
    \resizebox{\columnwidth}{!}{%
    \begin{tabular}{l|lll|lll}
    \toprule
     \multirow{2}{*}{Model}  & \multicolumn{3}{c}{Article-level} & \multicolumn{3}{c}{Quantity-level} \\
     & Type & Cond & Dir & Type & Ind & $+/-$ \\
     \midrule
     \textbf{Random} & 0.136 & 0.148 & 0.19 & 0.111 & 0.064 & 0.324 \\
    \textbf{Majority Label} & 0.237 & 0.269 & 0.252 & 0.193 & 0.08 & 0.382 \\
    \textbf{Mistral 0-shot} & 0.386 & 0.236 & 0.346 & 0.566 & 0.262 & 0.382 \\
    \textbf{Mistral 2-shot} & 0.367 & 0.186 & 0.385 & 0.358 & 0.47 & 0.43 \\
    \textbf{Base Classifier} & \textbf{0.515} & 0.697 & 0.493 & 0.685 & 0.824 & 0.796 \\
    \textbf{Base Classifier + DAPT} & 0.474 & 0.636 & 0.475 & 0.731 & 0.826 & 0.812 \\
    \textbf{Relational (best)} & 0.438 & \textbf{0.717} & \textbf{0.522} & \textbf{0.748} & \textbf{0.849} & \textbf{0.813}\\ 
     \bottomrule
    \end{tabular}}
    \caption{Avg. Macro F1 after 5-Fold Cross Validation}
    \label{tab:general_macro}
\end{table}

In Tab.~\ref{tab:psl_ablation} we include an ablation study for the relational model rules. We observe that all rules contribute to an increase in performance for their corresponding decision predicates (see bold scores). We find that combining hard constraints ($r1$,$r2$) and the rule modeling sequential dependencies ($r5$) performs the best. 

\begin{table}[t]
    \centering
    \resizebox{\columnwidth}{!}{%
    \begin{tabular}{l|lll|lll}
    \toprule
     \multirow{2}{*}{Rule}  & \multicolumn{3}{c}{Article-level} & \multicolumn{3}{c}{Quantity-level} \\
     & Type & Cond & Dir & Type & Ind & $+/-$ \\
     \midrule
     Priors only & \textbf{0.515} & 0.697 & 0.493 & 0.731 & 0.826 & 0.812 \\
$r1$ & 0.515 & 0.697 & 0.493 & 0.688 & 0.853 & \textbf{0.813} \\
$r2$ & 0.438 & \textbf{0.717} & 0.518 & 0.731 & 0.826 & 0.812 \\
$r3$ & 0.515 & 0.707 & 0.493 & 0.731 & 0.826 & 0.812 \\
$r4$ & 0.515 & 0.697 & 0.521 & 0.731 & 0.826 & 0.813 \\
$r5$ & 0.515 & 0.697 & 0.493 & 0.737 & 0.826 & 0.813 \\
$r1+r2$ & 0.438 & 0.717 & 0.518 & 0.688 & 0.853 & 0.813 \\
$r1 + r2 + r3$ & 0.433 & 0.7 & 0.493 & 0.697 & 0.854 & 0.813 \\
$r1 + r2 + r4$ & 0.432 & 0.691 & 0.516 & 0.697 & \textbf{0.855} & 0.813 \\
$r1 + r2 + r5$ \textbf{(best)} & 0.438 & \textbf{0.717} & \textbf{0.522} & \textbf{0.748} & 0.849 & \textbf{0.813} \\
$r1 + r2 + r3 +r4 +r5$ & 0.434 & 0.701 & 0.519 & 0.746 & 0.84 & 0.808\\
     
     \bottomrule
     
    \end{tabular}}
    \caption{Ablation (Macro F1) for the Reln. Model}
    \label{tab:psl_ablation}
\end{table}


We include results for the best relational model by article publisher in Tab.~\ref{tab:results_by_publisher}. At the article level, we observe higher values for the New York Times, reflecting the higher presence of annotated examples for this publisher (see Tab. ~\ref{tab:data_stats}). The results for Fox News articles are lower, reflecting the lower number of examples. Interestingly, the results for Breitbart and Huffpost do not follow this trend, maintaining relatively high values in spite of less supervision. In the case of quantities, we see relatively high and stable performance for all publishers, with the exception of Fox News.  We provide further error analysis in regard to this in App. ~\ref{app:error_analysis}. These results are particularly encouraging, as they suggest that we can leverage quantity predictions to perform a high-fidelity selection and framing analysis of economic news.

\begin{table}[t]
    \centering
    \resizebox{\columnwidth}{!}{%
    \begin{tabular}{l|lll|lll}
    \toprule
     \multirow{2}{*}{Publisher}  & \multicolumn{3}{c}{Article-level} & \multicolumn{3}{c}{Quantity-level} \\
     & Type & Cond & Dir & Type & Ind & $+/-$ \\
      \midrule
     New York Times & 0.544 & 0.809 & 0.534 & 0.663 & 0.826 & 0.777\\
     Wall Street Journal &  0.173 & 0.434 & 0.28 & 0.834 & 0.813 &  0.884 \\
     Washington Post & 0.574 & 0.704 & 0.621 & 0.665 & 0.686 & 0.788\\
     Fox News & 0.4 & 1.0 & 0.4 & 0.733 & 1.0 & 0.333\\
     Breitbart & 0.636 & 0.325 & 0.597 & 0.98 & 0.867 & 0.655\\  
     HuffPost & 0.564 & 0.625 & 0.3 & 0.609 & 0.686 &  0.866\\
    \bottomrule
    \end{tabular}}
    \caption{Macro F1 per Publisher with Best Reln. Model}
    \label{tab:results_by_publisher}
\end{table}

Lastly, we include fine-grained results for predicting the different macro-indicators using our best model in Tab.~\ref{tab:macro_results}. We find that we have relatively good performance for most indicators, with the exception of the 'other' category, which may be due to the lack of cohesiveness in this class. We also see lower performance for retail sales, which we attribute to the lack of sufficient supervision for this class (See Fig.~\ref{fig:label_dist_quant}). An effective large-scale analysis of retail indicators in the news would require greater supervision for this class. 

\begin{table}[t]
    \centering
    \resizebox{\columnwidth}{!}{
    \begin{tabular}{llll}
    \toprule
        Macro-Indicator & Prec. & Recall & F1 \\
    \midrule
       Job Numbers (jobs, unemployment) & 0.956 & 0.982 & 0.968 \\
       Retail Sales & 0.667 & 1.0 & 0.8 \\
       Interest Rates (Fed, mortgage)& 0.75 & 0.857 & 0.8 \\
       Prices (CPI, PPI) & 0.733 & 0.846 & 0.786 \\
       Energy Prices (gas, oil, etc.) & 1.0 & 1.0 & 1.0 \\
       Wages & 0.867 & 0.812 & 0.839 \\
       Macro Economy (GDP, etc) & 0.881 & 0.860 & 0.871 \\
       Market Numbers (any financial market) & 0.941 & 0.814 & 0.873 \\
       Currency Values & 1.0 & 0.8 & 0.889 \\
       Housing (start, sales, pricing) & 1.0 & 0.95 & 0.974 \\
       Other & 0.375 & 0.5 & 0.429 \\
       None & 0.956 & 0.964 & 0.96 \\
       \midrule 
       \textbf{Accuracy} & 0.922 & 0.922 & 0.922 \\
       \textbf{Macro Average} & 0.844 & 0.865 & 0.849 \\
       \textbf{Weighted Average} & 0.926 & 0.922 & 0.923\\
       \bottomrule
    \end{tabular}}
    \caption{Macro-Indicator results with Best Reln. Model}
    \label{tab:macro_results}
\end{table}

\subsection{Framing and Selection Analysis}\label{sec:analysis}

In this section we present a brief exploration of how the methods outlined in this paper can be used to study editorial choices. To do this, we use our best model to predict all frame components for the full dataset of 199,066 articles. We then use these frames to show how economic messaging shifts in response to exogenous shocks and changes in the balance of political power. In particular, we show how our framework can help us understand how the different choices these major publications make can produce significantly different views of economic conditions.

\paragraph{Article-level framing with macro-indicators} Because we have greater confidence in our model's quantity-level performance, we generate article-level frames for specific economic indicators using our quantity-level predictions. First, we select the subset of articles that include at least two quantities in the category of interest. Then, we assign a positive indicator frame to articles with at least twice as many quantities with a positive polarity than negative polarity and vice versa for negative frames. Articles that meet neither threshold are assigned a "neutral" frame.

\begin{figure}[t]
    \centering
    \includegraphics[width=\columnwidth]{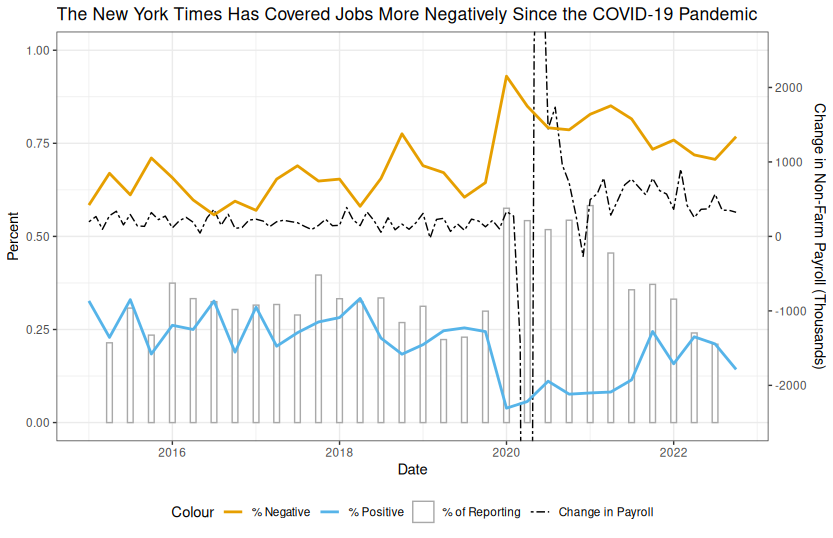}
    \caption{Framing of articles referencing job numbers from 2015 through 2023 in the New York Times.\footnote{The Y-axis has been limited to account for the massive swing in employment during the early pandemic.} Spin was aggregated quarterly. Monthly payroll (employment) data can be seen in the dotted black line. The gray bars represent proportion of the overall coverage taken up by jobs reporting.}
    \label{fig:jobs}
\end{figure}

\paragraph{Shifts in Framing} Our preliminary evidence suggests that the framing of articles on any given topic by any given publication is both sensitive to exogenous factors as well as decisions made by the publication itself. Fig. \ref{fig:jobs} illustrates how the \textit{New York Times (NYT)} framed their articles in regard to jobs data from 2015 to 2023. First, we can see that prior to 2020, during a period of stable job growth, the \textit{NYT} already had a sustained negative valance, matching the key findings in recent literature on the general negativity bias on the Mainstream Media \cite{NBERw32026}. In response to the onset of COVID in 2020, the \textit{NYT} intensified its coverage of the job market in response to a dramatic surge in unemployment caused by a cascade of lockdowns. We can also see that much of this coverage carried an extreme negative valence, which matches the large drop in jobs and subsequent uncertainty. 

As the job market recovered and the labor market entered an extended period of exceptionally high growth, the volume of reporting in the \textit{NYT} decreased back to its pre-pandemic levels, but the negativity persisted. This was not the same for other publications such as the Wall Street Journal, whose coverage more closely resembles the shifts in valence in the supporting data (See App. \ref{app:analysis_graphs}.)

\begin{figure}[t]
    \centering
    \includegraphics[width=\columnwidth]{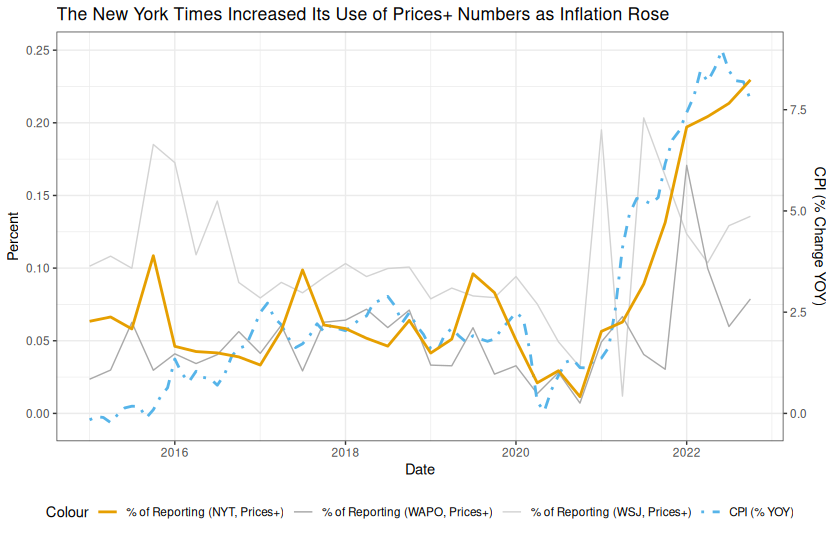}
    \caption{Selection of economic indicators referencing price (price \& energy) numbers from 2015 through 2023 in the New York Times, Washington Post, and Wall Street Journal. Aggregated Quarterly. Monthly CPI data can be seen in the dotted blue line.}
    \label{fig:prices}
\end{figure}



\paragraph{Indicator Selection} Fig. \ref{fig:prices} shows us another lever that publications can pull when shaping their coverage: indicator selection. As inflation began to rise, we can see that all three publications responded by increasing their coverage of prices through a greater use of price based indicators.\footnote{This includes energy (e.g. oil) prices.} These three publications did not, however, respond in an identical manner, with, for example, the \textit{New York Times} increasing their coverage of prices far after the other publications returned to pre-crisis levels of coverage.

With the simple analysis shown in these two figures, we can begin to show that editors have considerable leeway when choosing what indicator to cover at any given time, and how to cover it. While we kept the analysis brief due to space considerations, we were able to show the usability of our framework to track a diverse set of indicators reported in different publishers across time.

\section{Applicability to Other Domains}
We emphasize that the main contribution of this paper is not centered around performance gains of our relational model over the baseline models, but rather the operationalization of a challenging task: analyzing framing at a scale for domains with supporting numerical data, such as the economy. While our solution does not apply to all framing scenarios, it applies to a substantial number. For example, we envision this framework being useful for analyzing news about crime (by tracking data points maintained by the FBI and local jurisdictions), climate change (by tracking weather statistics, global avg. temperature, emissions statistics, etc.) and public-opinion (by tracking polling numbers, political betting market numbers, etc.).



\section{Conclusion and Future Work}

In this paper, we presented a novel framework for modeling framing in news articles in the presence of supporting data. We focused our analysis on news about the U.S. economy, and proposed an annotation schema to identify framing at three levels of abstraction: the general framing of the economic conditions, the economic indicators that are highlighted, and the framing of each of those indicators.

We showed that we can predict the components of our schema with relatively good performance, even in very low supervision settings. Finally, we demonstrated how our framework can be used to perform a large scale exploration of the frame choices for a given economic indicator. 

In our analysis, we showed that different publications cover economics using different indicators, for example the \emph{NYT} covers jobs numbers less than normal during a strong jobs recovery, instead switching to more coverage rising prices. And, while all three publications responded to rising inflation, they each used price numbers differently when covering the economy, with the \emph{NYT} employing price numbers more frequently than the \emph{Washington Post}, or the \emph{Wall Street Journal}.


Going forward, we have two main points of focus. 
First, we are working on expanding our schema to capture a larger set of economic aspects. While we focused our analysis on macro-economic indicators, there are other indicators that we could track to get a more holistic view of the economy. For example, for governmental indicators, we could track types of expenditures (e.g. social security), revenue (e.g. taxes), as well as debt and deficit. 
Additionally, we want to perform a large-scale analysis of economic news framing. In this paper, we presented a brief demonstration of how our framework can be used to analyze shifts in framing across time for different publications. Next, we want to perform a similar analysis for a larger set of economic indicators and publications. We note that to perform such an analysis we must scale up our annotations and improve our model predictions.


\section{Limitations}

The work presented in this paper has two main limitations:

(1) Obtaining high-quality annotations for our frame structure is very expensive and time consuming. For this reason, we worked with a low amount of supervision. While we showed that we could obtain relatively good performance in this constrained scenario, to be able to use our framework for a large scale, holistic analysis, we need to address this issue. This is particularly important given that our annotations are considerably skewed. For example, we have significantly more supervision for job numbers and market numbers (which occur more often) than we do for energy prices (which occur less often). This considerably affects our performance for more long-tail indicators, and hence limits the types of analysis that we can perform. Our current efforts are dedicated to increase the amount of annotations for all components of our frame structure, as well as incorporating alternative semi-supervised strategies. We are hoping to release an extended version of the annotated dataset to the community in the near future. 

(2) The fact that we are using automated techniques for frame prediction necessarily carries some uncertainty. Even if we improve our models considerably, our large scale analysis will have a margin of error. It is important to acknowledge this when presenting our findings.  
\section{Ethical Considerations}

To the best of our knowledge, no code of ethics was violated during the development of this project. We used publicly available tools to collect our dataset, and discarded any instances that were unreachable. The annotators were paid \$15 per hour, and no personally identifiable information was collected or recorded during annotation. 

We performed a thorough evaluation of our dataset, which is presented in the
paper. We reported all pre-processing steps, learning configurations, hyperparameters, and additional technical details. Due to space constraints, some of this information was relegated to the Appendix. Further, the data and code have been released to the community. The results reported in this paper support our claims and we believe that they are reproducible. 

The analysis reported in Section \ref{sec:analysis} is done using the outputs of a machine learning model and does not represent the authors personal views. The uncertainty of these predictions was adequately acknowledged in the Limitations Section, and the estimated accuracy was reported in Section \ref{sec:prediction}.



\section{Acknowledgements}

This work utilized the Alpine high-performance computing resource, the Blanca condo computing resource, and the CUmulus on-premise cloud service at the University of Colorado Boulder. Alpine is jointly funded by the University of Colorado Boulder, the University of Colorado Anschutz, Colorado State University, and the NSF (award 2201538). Blanca is jointly funded by computing users and the University of Colorado Boulder. CUmulus is jointly funded by the National Science Foundation (award OAC-1925766) and the University of Colorado Boulder.

\bibliography{anthology,custom}
\bibliographystyle{acl_natbib}

\appendix

\section{Appendix}

\subsection{Additional Data Collection Details}\label{app:data_collection}

For each of our included publishers' landing pages, we retrieved a list of each Wayback Machine entry from Jan. 1st 2015 to Jan. 1st 2023, segmented the list of entries by the hour, and downloaded the earliest available document in that hour. Each of these documents is considered ``front page''.

We used a heuristic based on the Document Object Model tree structure to remove opinion pieces, identify news articles, and establish an apparent ranking of those articles based on rendered size and distance from the upper-left hand corner of the viewport. 
This was calculated as a proxy for the order in which a reader sees the headlines. 

Each article is assumed to have one and only one hypertext reference (href) across time, which points to a URL for the article's HTML document. We recorded the front page document's scrape timestamp, and each article's href and apparent rank. For each unique article-timestamp tuple, we retrieved the linked document from the Wayback Machine.
The timestamp is unlikely to match exactly, so we retrieved the nearest available entry. We extracted plain text versions of the article document whenever it was available, using Trafilatura \cite{barbaresi-2021-trafilatura}, and Readability \cite{ReadabilityJs2023, ReadabiliPy2023}. These are both software packages designed to extract only the text from an HTML document. Trafilatura is built with natural language processing techniques in mind, whereas Readability is designed for naturalistic interfaces, specifically the "reader" view in Mozilla Firefox.

\begin{figure}[ht]
\includegraphics[width=\columnwidth]{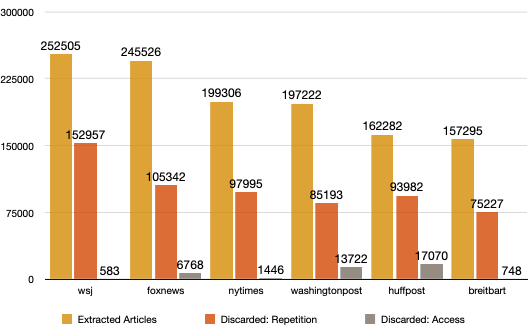}
\caption{Discarded Articles per Source}\label{fig:discarded}
\end{figure}

In some cases, the main body of the article could not be extracted either because the link no longer exists, or because the content is behind a paywall. We removed all such instances. We discarded all duplicates, keeping only the first appearance of each article. Fig.~\ref{fig:discarded} shows the number of articles that were discarded due to either repetition or access issues. The number of discarded articles is quite large, but given that our dataset is based on hourly snapshots for the archive, this is to be expected. We aimed to over collect during the early collection steps to maximize the completeness of our final dataset.

\begin{table}[ht]
\begin{adjustbox}{width=\columnwidth, center}
\begin{tabular}{|l|}
\hline
bond*, budget, business, consumer, consumer price index, \\
cost, cpi, currency debt, deficit, demand, dow jones, earning*, \\
econo*, employ*, expenditure*, export*, fed, financ*, fiscal, gdp, \\
gross domestic product, housing, import*, income,  ipi, ipp, \\
industrial production index, inflation, interest, invest*, \\
international price program, job*, labor, market, monet*, mortgage, \\
nasdaq, poverty, ppi, price*, productivity, producer price index, \\
retail, revenue, s\&p 500, sales, securities, small cap 2000, stimulus,\\
 stock, supply,
tax, trade, trading, treasur*, unemploy*, wage*,  wti, \\
west texas intermediate \\
     \hline
    \end{tabular}%
\end{adjustbox}
\caption{Economic Lexicon used to determine whether a given article pertains to the economy.} 
\label{tab:keywords}
\end{table}

We stored the document's title as extracted by readability and regarded it as the article's headline. In some cases, there were errors when extracting headlines. To deal with these cases, we used T5~\cite{10.5555/3455716.3455856}, a generative language model, to generate headlines using the body of the article. 
To do this, we used a model that was trained on 500,000 articles with their headlines~\footnote{huggingface.co/Michau/t5-base-en-generate-headline}.


To identify the subset of articles that discuss the economy, we curated a lexicon of economic terms, shown in Table~\ref{tab:keywords}. We sourced these terms from the Bureau of Labor Statistics and the FRED database operated by the Federal Reserve bank of St. Louis.\footnote{https://www.bls.gov/data/}\footnote{https://fred.stlouisfed.org/} These sources provide a fairly comprehensive list of common metrics of economic activity.

\begin{figure}[t]
\centering
\includegraphics[width=\columnwidth]{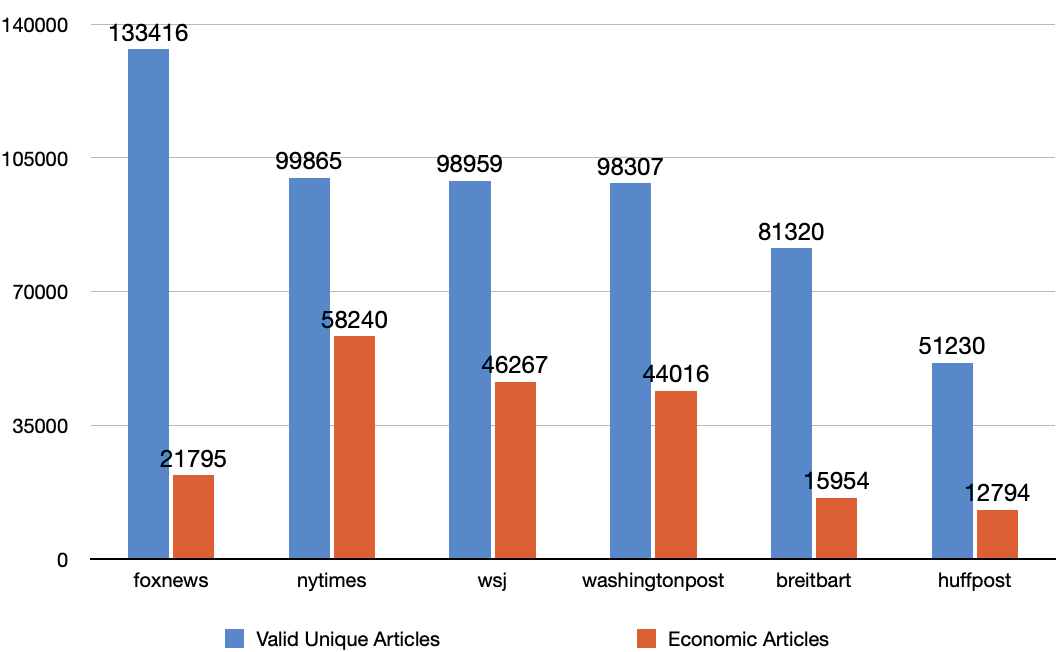}
\caption{Economic Articles per Publisher}
\label{fig:extracted_stats}
\end{figure}

We consider an article to be relevant to the economy if it contains at least three sentences mentioning any economic term from our lexicon. This resulted in a total of 199,066 articles (See Table~\ref{tab:data_stats}). Fig.~\ref{fig:extracted_stats} shows the number of valid, unique articles extracted per source, as well as the resulting number of economic articles.

\subsection{Codebook Development}\label{app:codebook_dev}
The codebook (included in Appendix \ref{app:ann_guidelines}) was developed by the two senior authors of the paper, one economist and one NLP/CSS expert.

\paragraph{Codifying indicators} We had an initial hypothesis that the reporting of economic indicators was prevalent in economic news, so we performed a few rounds of manual analysis using standard qualitative practices to establish the relevance of this information in our large unlabeled dataset. During this process, we realized that there were many relevant mentions to indicators that were not macro-economic, and we introduced the “type” category to differentiate between them (e.g., personal, industry, government). We created the initial lists of subcategories based on standard reported indicators, and refined them after a few rounds of annotation, taking annotator feedback into account.

\paragraph{Codifying article-level categories} We inherited the type category from the indicators to codify the “prevalent” or “dominant” type of information that the article discussed. For the economic conditions and direction, we took the Gallup economic index questions\footnote{https://news.gallup.com/poll/1609/consumer-views-economy.aspx}, and re-worded them to reflect the fact that coders were reading an article from sometime in the last 8 years. Finally, we refined them after a few rounds of annotation, taking annotator feedback into account.

\subsection{Annotation Guidelines}\label{app:ann_guidelines}

In the following sections we replicate the codebook provided to the coders. This codebook was used during training, and as a reference for the coders to use as needed. 
\subsubsection{Setup and Goals}
We have a large set of news articles about general economic topics. The goal is to: 
\begin{enumerate}
    \item{Identify if the article makes reference to macroeconomic conditions in the US. If it does then, answer the following questions: }
    \begin{itemize}
        \item How does this article rate economic conditions in the US today -- (Good; Poor; No Opinion)
        \item Does this article state/imply that economic conditions in the US as a whole are…? (Getting better; Getting worse; Same; No Opinion)
    \end{itemize}
    \item If it does not, we would like to identify what the article covers: industry-specific economic information, firm-specific economic information, government-specific or political information, or personal stories. 
    \item  In a given article, we want to look at quantities and figures being reported and identify the type of figure (i.e. macro-economic, industry-specific, firm-specific, personal story), and in case of a macroeconomic, industry or government figure, identify the specific sub-type/indicator being discussed. 

\end{enumerate}

\subsubsection{Guidelines}

To annotate specific ARTICLES we will:

\begin{enumerate}
    \item Click “Edit Article” 
    \item Annotate key quantities (first 5 relevant quantities in an article + 2 scattered throughout the article, if present):
    \begin{itemize}
        \item Each quantity identified will be highlighted in yellow and clickable
        \item If the quantity is a date or otherwise not relevant, skip it
        \item If the quantity is relevant, click it and follow the instructions (code-book will be outlined below)
        \item Data points may not fit into one of the groups. These should be labeled "Other" and a short comment should be left to explain it
        \item Annotate general information: Answer the general article questions (detailed code-book below) 
    \end{itemize}
\end{enumerate}

\subsubsection{Codebook}

\begin{enumerate}
    \item \textbf{Macro-economic}
    \begin{itemize}
        \item Macro-economic indicators aggregate data according to a set of demographics.
        \item Data Guide
        \begin{itemize}
            \item For the purpose of this study, these will mostly be values that are aggregated up to a national level (e.g. national GDP). There may be, however, other demographic groups, like states, that fall into this category
        \end{itemize}
   
        \item Types of macro-indicators:
        \begin{itemize}
            \item Jobs Numbers (Jobs, Unemployment)
            \item Market Numbers (any financial market)
            \item Housing (Start, Sales, Pricing)
            \item Macro Economy (GDP, etc.)
            \item Wages
            \item Prices (CPI, PPI)
            \item Confidence
            \item Retail Sales
            \item Interest Rates (Fed., Mortgage)
            \item Currency Values
            \item Energy Prices (Gas, Oil, etc.)
            \item Other (Specify)
        \end{itemize}
    \end{itemize}
    \item \textbf{Firm-Specific}
    \begin{itemize}
        \item A firm-specific data point is a data point associated with a particular firm or company. 
        \item Data Guide:
        \begin{itemize}
            \item Examples include: stock prices, debt offerings, and capital investments
        \end{itemize}  
    \end{itemize}
    \item \textbf{Industry-Specific}
    \begin{itemize}
        \item Industry level articles/quantities describe an entire industry rather than individual businesses
        \item Data Guide:
        \begin{itemize}
            \item Examples include: Chip Manufacturers Post Strong Growth Numbers
        \end{itemize}
        \item Types of Industries:
        \begin{itemize}
            \item Agriculture, forestry and hunting
            \item Mining
            \item Utilities
            \item Construction
            \item Manufacturing
            \item Wholesale trade - \textit{selling products in bulk to other businesses}
            \item Retail trade - \textit{selling products directly to the end consumer}
            \item Transportation and warehousing
            \item Information
            \item Finance, insurance, real estate, rental and leasing
            \item Professional and business services
            \item Educational services, health care, and social assistance
            \item Arts, entertainment, recreation, accommodation and food services
            \item Other (Except government)
        \end{itemize}
    \end{itemize}
    \item \textbf{Government Revenue and Expenditures}
    \begin{itemize}
        \item Any value that describes how a government earned or spent its income falls into this category. 
        \item Data Guide:
        \begin{itemize}
            \item A few examples are: taxes, budgets, and treasury issuances
        \end{itemize}
        \item Government Level:
        \begin{itemize}
            \item Federal
            \item State and local
        \end{itemize}
        \item Types of Expenditures
        \begin{itemize}
            \item Social Security and Public Welfare
            \item Health and Hospitals
            \item National Defense
            \item Police
            \item Transportation
            \item Research
            \item Education
            \item Employment
            \item Housing
            \item Corrections
            \item Courts 
            \item Net Interest 
            \item Other
        \end{itemize}
        \item Types of Revenue 
        \begin{itemize}
            \item Taxes and other compulsory transfers imposed by government units
Property income derived from the ownership of assets
            \item Sales of goods and services
            \item Voluntary transfers received from other units
        \end{itemize}
    \end{itemize}
    \item \textbf{Personal}
    \begin{itemize}
        \item If an article/quantity focuses on the economic condition of a single person, or a group of individuals that is not large enough to represent an entire demographic then we consider it personal. E.g. an individual's struggle to find work, or the grocery/gas budget of a single family
        \item Data Guide:
        \begin{itemize}
            \item E.g. household expenditures, personal debts
        \end{itemize}
        \item A short comment should be added to explain the frame/quantity
    \end{itemize}
\end{enumerate}

\subsubsection{Additional Notes}

\textbf{Data Points:}

\begin{itemize}
    \item Each candidate data point is highlighted and clickable
    \item Each data point has a general type: 
    \begin{itemize}
        \item Macro-economic:
        \item Firm-specific
        \item Industry-specific
        \item Government revenue and expenditures
        \item Personal
    \end{itemize}
\end{itemize}

\textbf{General Frame:}
\begin{itemize}
    \item While an article can reference different data points, the goal is to identify the dominant frame
    \item An article can have only one dominant frame
    \begin{itemize}
        \item Macro-economic 
        \item Firm-specific
        \item Industry-specific
        \item Government revenue and expenditures
        \item Personal 
        \item Other
    \end{itemize}
    \item An article may frame the economy in a certain light – if its dominant frame is Macro-economic, it will be tied to the following two questions:
    \begin{itemize}
        \item How \textbf{does this article} rate economic conditions in the US -- (Excellent; Good; Only Fair; Poor; No Opinion; Not relevant to the US economy)
        \item \textbf{Does this article} state/imply that economic conditions in the US as a whole are…? \{Getting better; Getting worse; Same; No Opinion; Not relevant to the US economy\}
    \end{itemize}
    \item A comment should be added to explain rationale for choosing the frame class / economic outlook
\end{itemize}

\subsection{Cross-Annotation Statistics}\label{app:cross_ann_stats}

Below in Table~\ref{tab:cross_ann_stats} we present the statistics for the number of people who cross-annotated each of the label categories.

\begin{table}[ht]
    \centering
    \resizebox{\columnwidth}{!}{%
    \begin{tabular}{l|ccc}
        \toprule
        \multicolumn{1}{c}{\textbf{}} & \multicolumn{3}{c}{\textbf{Instances annotated by 2, 3 and 4+ annotators}} \\
        \midrule
        \textbf{Annotation} & \textbf{2 Annotators} & \textbf{3 Annotators} & \textbf{4+ Annotators} \\
        \midrule
        Article Type & 17.19\% & 49.22\% & 33.59\% \\
        Econ. Conditions & 49.22\% & 55.13\% & 17.95\% \\
        Econ. Direction & 26.92\% & 55.13\% & 17.95\% \\
        Quantity Type & 55.59\% & 35.85\% & 8.56\% \\
        Macro Ind. & 51.6\% & 41.44\% & 0.36\% \\
        Quantity Polarity & 78.46\% & 21.18\% & 6.95\% \\
        \bottomrule
    \end{tabular}}
    \caption{Percentage of annotation instances annotated by 2, 3, and 4+ annotators}
    \label{tab:cross_ann_stats}
\end{table}

\subsection{Annotation Confusion Matrices}\label{app:cm_ann}

To further characterize agreement values and types of disagreements for all annotations, we provide a set of confusion matrices below. We also provide specific examples of ambiguous cases.

\begin{figure}[!htb]
    \centering
    \includegraphics[width=\columnwidth]{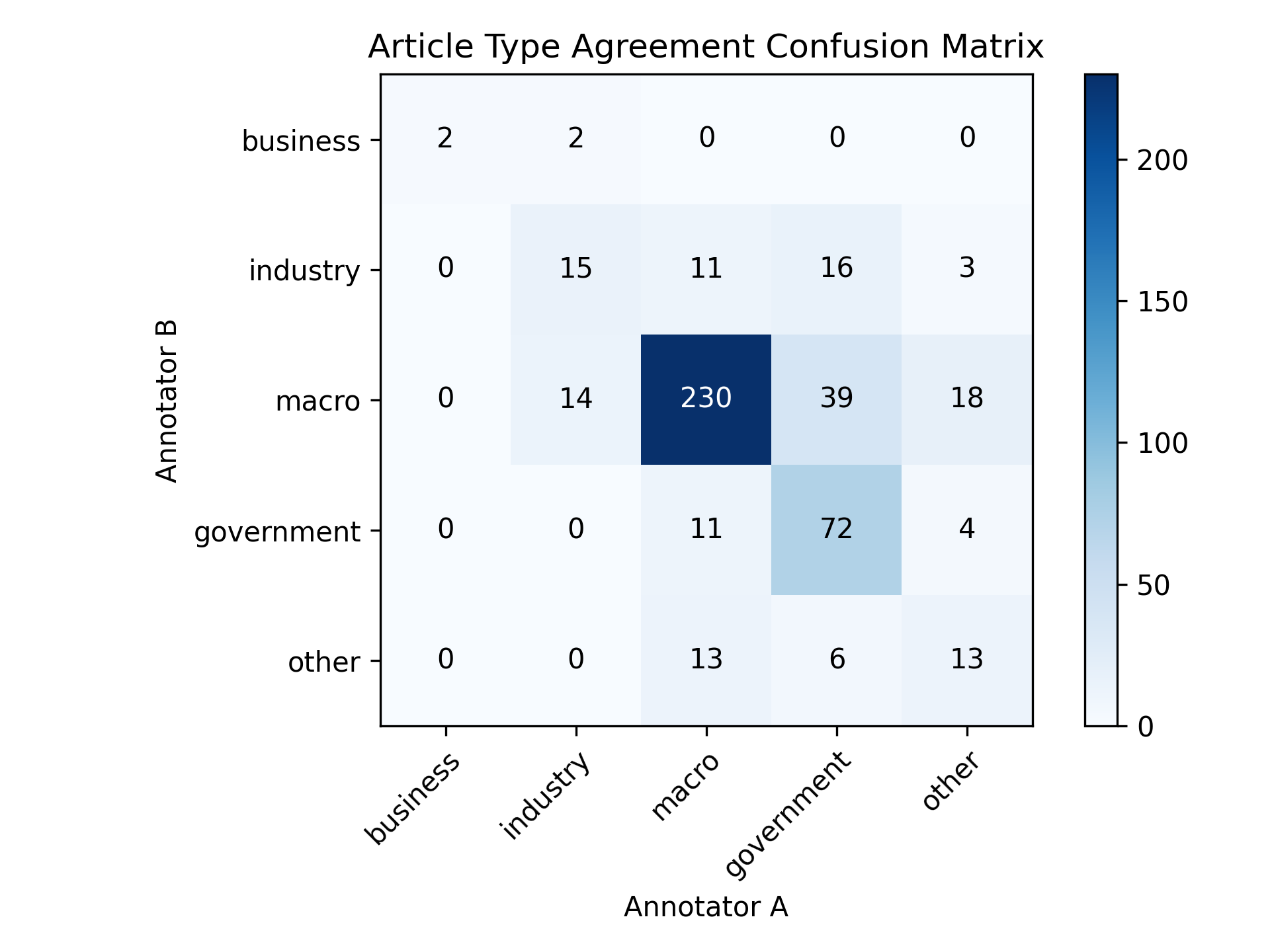}
    \caption{Types of agreement and disagreement among annotators for Article Type}
    \label{fig:cm_frame}
\end{figure}

We see in Fig.~\ref{fig:cm_frame} that in general, there is some confusion between \textit{macro} and \textit{government} types. This type of disagreement was seen with The New York Times article entitled \href{https://www.nytimes.com/2018/08/01/us/politics/trump-short-term-health-insurance.html}{“Short Term’ Health Insurance? Up to 3 Years Under New Trump Policy”}, which was classified with an article type \textit{macro} and \textit{government} by different annotators. The article focuses heavily on the policies of the Trump administration, which falls under \textit{government}. However, it also makes strong arguments regarding rising prices, which falls under \textit{macro}.

\begin{figure}[!htb]
    \centering
    \includegraphics[width=\columnwidth]{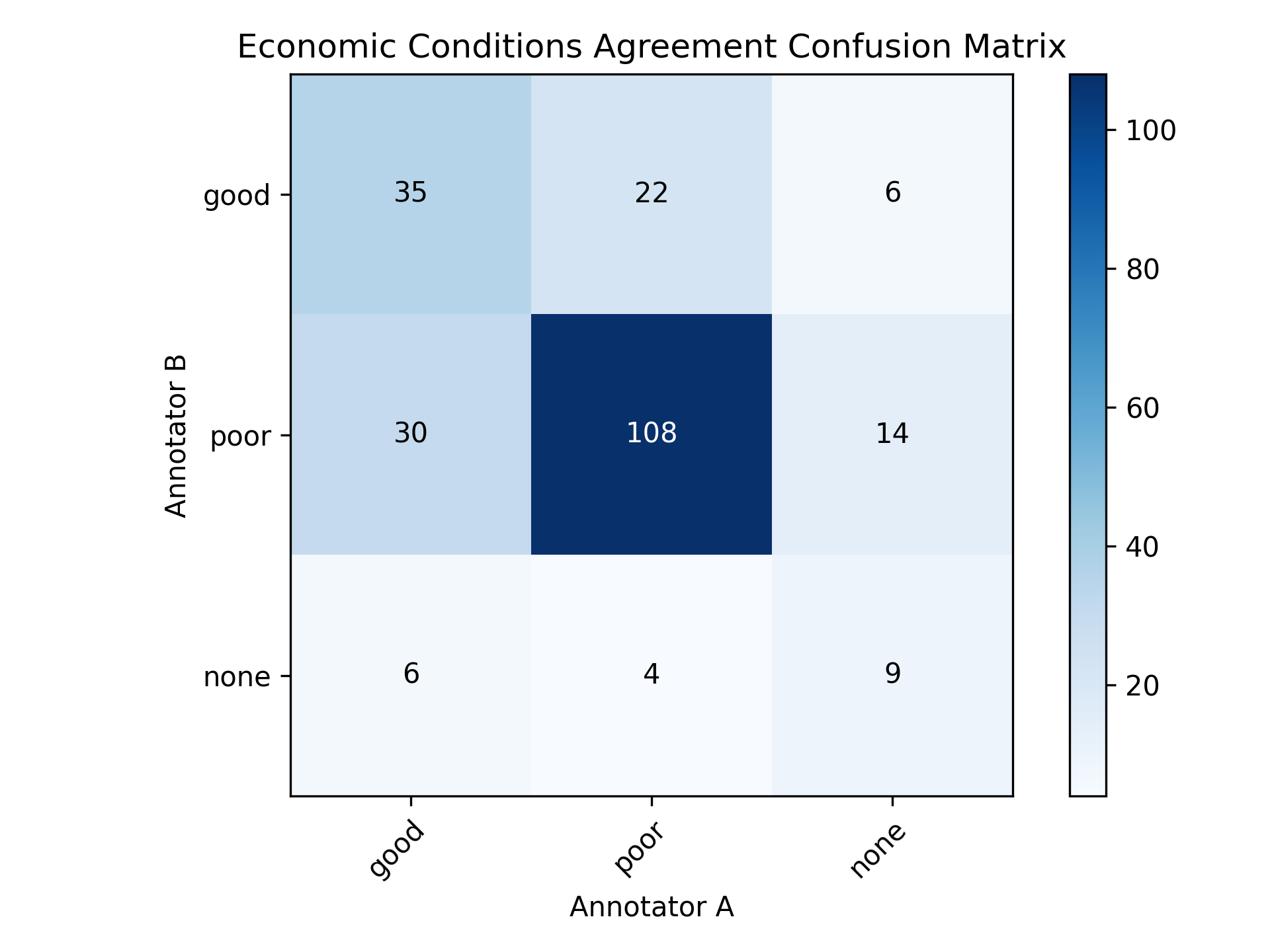}
    \caption{Types of agreement and disagreement among annotators for article-level Economic Condition}
    \label{fig:cm_econ_rate}
\end{figure}

In general, we see in Fig.~\ref{fig:cm_econ_rate} there is strong agreement when the article is negative towards the economy. However, annotators struggle more to differentiate positive articles. One example of this type of disagreement is The Wall Street Journal article entitled  \href{https://www.wsj.com/video/fed-faces-mixed-signals-as-hiring-cooled-in-august/A61504D6-6783-4C93-B1A2-87C0E1D21743}{“Fed Faces Mixed Signals as Hiring Cooled in August"}, which was classified as framing economic conditions as both \textit{good} and \textit{poor} by different annotators. We can observe that even the title of this article is inherently ambiguous, containing "mixed signals". 

\begin{figure}[!htb]
    \centering
    \includegraphics[width=\columnwidth]{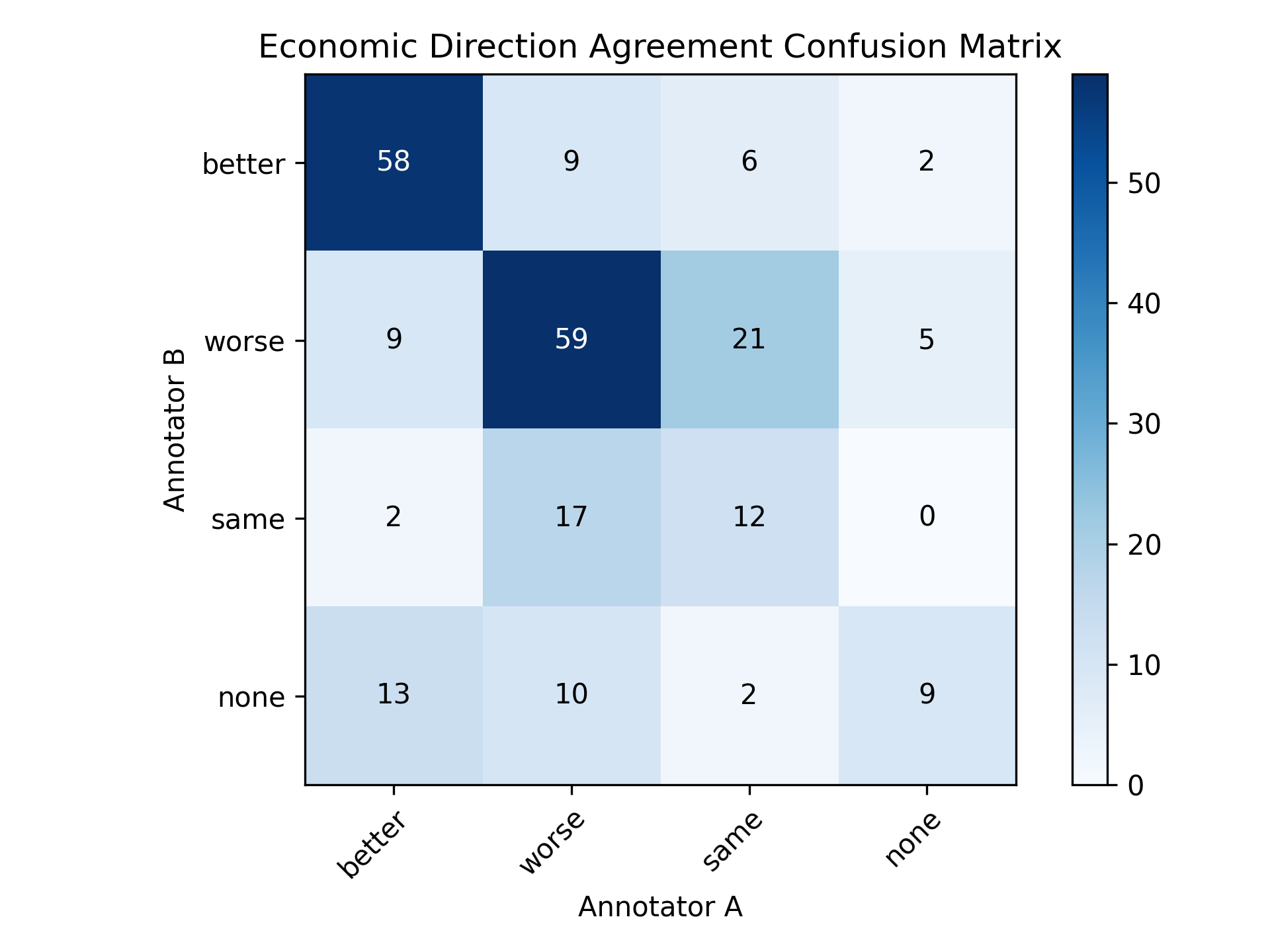}
    \caption{Types of agreement and disagreement among annotators for article-level Economic Condition}
    \label{fig:cm_econ_change}
\end{figure}

 We see in Fig.~\ref{fig:cm_econ_change} that there is strong agreement when an article is positive or negative towards the future of the economy. However, there is more confusion w.r.t. the \textit{same} category. One example of this is seen with the Huffington Post article  \href{https://imgur.com/a/qgjKfvo}{“The Secret IRS Files: Trove Of Never-Before-Seen Records Reveal How The Wealthiest Avoid Income Tax”}, which was classified as framing economic direction as both \textit{worse} and \textit{same}. The article mentions rising wealth inequality (which could align with \textit{worse}) but that does not really center the topic or talk explicitly about any forecast (which could align with \textit{same}).

\begin{figure}[!htb]
    \centering
    \includegraphics[width=\columnwidth]{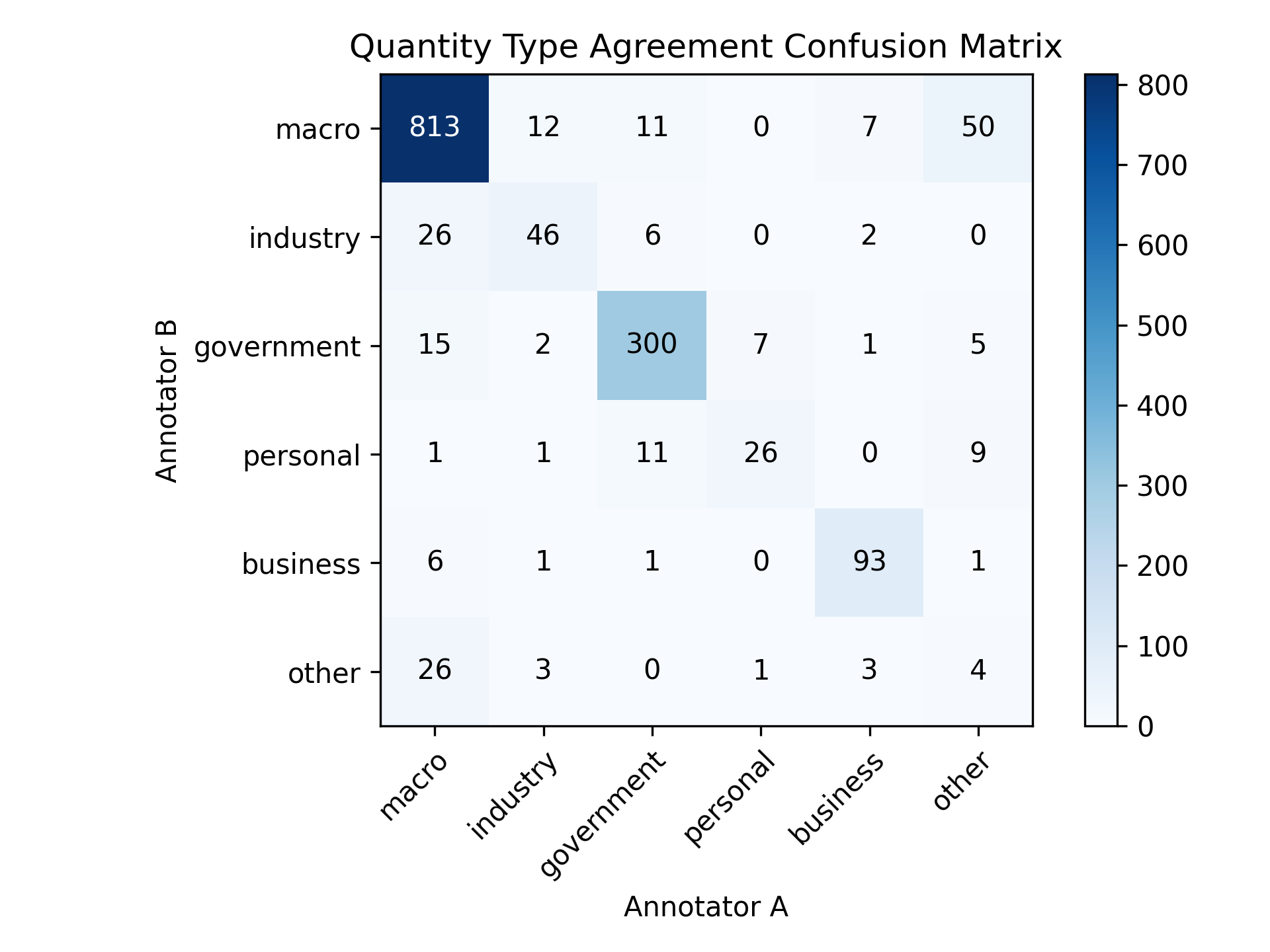}
    \caption{Types of agreement and disagreement among annotators for Quantity Type}
    \label{fig:cm_type}
\end{figure}

In Fig.~\ref{fig:cm_type} we observe a significant number of disagreements between \textit{industry} and \textit{macro} values for Quantity Type annotations. One example is seen in the excerpt: "Daily Business Briefing Oil prices approached \textbf{\$100 a barrel} on Tuesday, the highest in more than seven years..." where the focus is on annotating the "\$100 a barrel" indicator. We acknowledge that oil prices are an industry-specific topic but also have an effect on the macro economy. As a result, this case is ambiguous. 

\begin{figure}[!htb]
    \centering
    \includegraphics[width=\columnwidth]{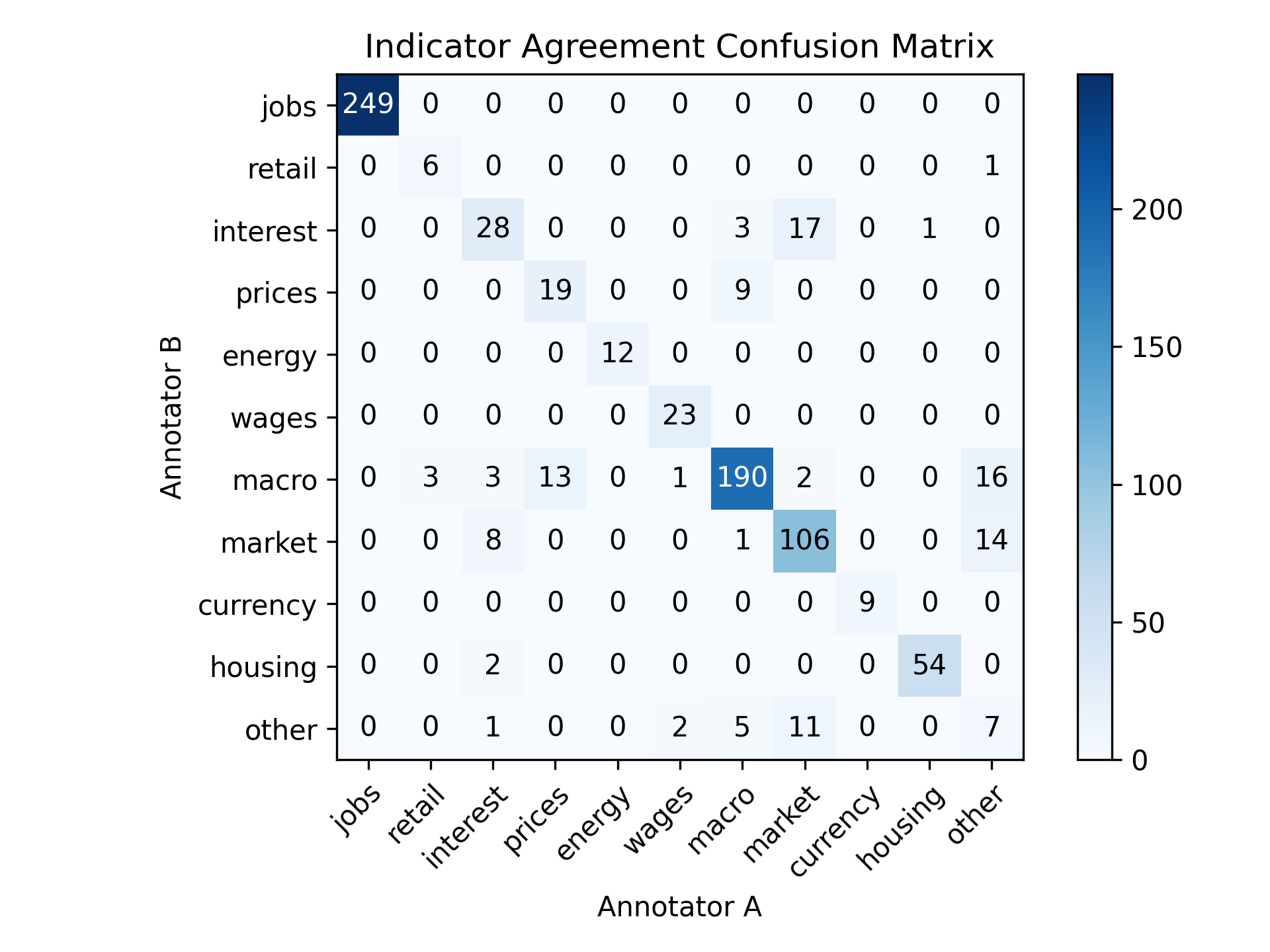}
    \caption{Types of agreement and disagreement among annotators for Indicator Type}
    \label{fig:cm_macro_type}
\end{figure}

In Fig.~\ref{fig:cm_macro_type} we see that annotators have difficulty discerning between \textit{prices} and \textit{macro} indicators. One ambiguous case is the excerpt: "But he said that declines in goods prices and rents, which have contributed notably to inflation over the last 18 months, might be insufficient if firms don’t slow their hiring. 'The labor market ... shows only tentative signs of rebalancing, and wage growth remains well above levels that would be consistent \textbf{with 2\% inflation},' Mr. Powell said." where the indicator in question is "with 2\% inflation". While the indicator is discussing inflation--which is a macro topic--it is also strongly related to prices.

\begin{figure}[!htb]
    \centering
    \includegraphics[width=\columnwidth]{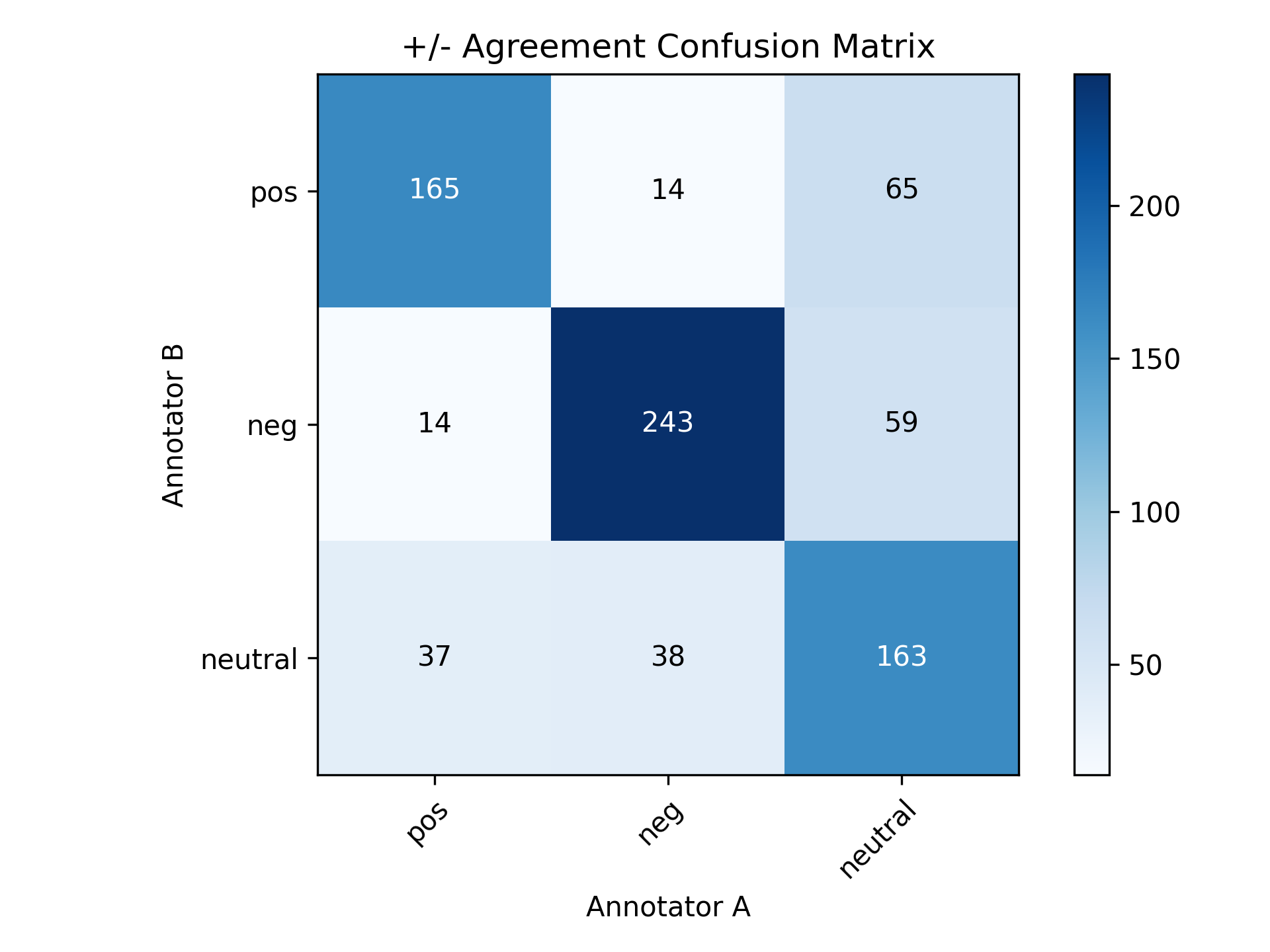}
    \caption{Types of agreement and disagreement among annotators for Indicator Polarity}
    \label{fig:cm_spin}
\end{figure}

In Fig.~\ref{fig:cm_spin} we see a high volume of disagreement between the \textit{pos} and \textit{neutral} labels. This type of disagreement was seen in the excerpt: "Economists expect applications for jobless benefits—seen as a proxy for layoffs—ticked down to 825,000 last week from 837,000 a week earlier. Weekly jobless claims are down sharply from a peak of nearly seven million in March but have clocked in at between \textbf{800,000} and 900,000 for more than a month. Claims remain above the pre-pandemic high of 695,000." where the indicator in question is "800,000". While the excerpt positively compares current joblessness positively to the previous week's numbers, but negatively to pre-pandemic numbers, making the value difficult to annotate.  

\subsection{Additional Details for Experimental Setup}\label{app:setup}
All experiments were run on a server with a single NVIDIA Tesla A100 GPU with 40 GB of RAM. In Table~\ref{tab:split_counts} we include the number of data points in each data split used for 5-fold cross-validation. All "Agreed" data points were cross-annotated by 2 or more people. Noisy data points include all annotations for labels which there was no inter-annotator agreement.

\begin{table}[ht]
    \centering
    \resizebox{\columnwidth}{!}{%
    \begin{tabular}{l|rrrrrr|rrrrrr}
        \toprule
        \multicolumn{1}{l}{} & \multicolumn{6}{c}{\textbf{Article Level}} & \multicolumn{6}{c}{\textbf{Quantity-level}} \\
        \midrule
        \multicolumn{1}{l}{} & \multicolumn{2}{c}{\textbf{Type}} & \multicolumn{2}{c}{\textbf{Condition}} & \multicolumn{2}{c}{\textbf{Direction}} & \multicolumn{2}{c}{\textbf{Type}} & \multicolumn{2}{c}{\textbf{Indicator}} & \multicolumn{2}{c}{\textbf{+/-}} \\
        \multicolumn{1}{l}{\textbf{Fold}} & \multicolumn{1}{l}{\textbf{Agreed}} & \multicolumn{1}{l}{\textbf{Noisy}} & \multicolumn{1}{l}{\textbf{Agreed}} & \multicolumn{1}{l}{\textbf{Noisy}} & \multicolumn{1}{l}{\textbf{Agreed}} & \multicolumn{1}{l}{\textbf{Noisy}} & \multicolumn{1}{l}{\textbf{Agreed}} & \multicolumn{1}{l}{\textbf{Noisy}} & \multicolumn{1}{l}{\textbf{Agreed}} & \multicolumn{1}{l}{\textbf{Noisy}} & \multicolumn{1}{l}{\textbf{Agreed}} & \multicolumn{1}{l}{\textbf{Noisy}} \\
        \midrule
        0 & 22 & 58 & 15 & 68 & 15 & 74 & 110 & 683 & 56 & 384 & 78 & 789 \\
        1 & 25 & 58 & 17 & 68 & 17 & 74 & 133 & 704 & 108 & 341 & 85 & 773 \\
        2 & 22 & 58 & 11 & 68 & 10 & 74 & 124 & 707 & 50 & 378 & 77 & 783 \\
        3 & 20 & 58 & 11 & 68 & 10 & 74 & 108 & 636 & 65 & 343 & 82 & 744 \\
        4 & 22 & 58 & 12 & 68 & 12 & 74 & 137 & 693 & 66 & 337 & 105 & 771 \\
        \bottomrule
    \end{tabular}}
    \caption{Number of examples in each data split used for 5-fold cross-validation}
    \label{tab:split_counts}
\end{table}

\paragraph{Mistral baseline details}
To further motivate our modeling choices, we present baseline results for zero and few-shot prompting instruction tuned model, Mistral-7B-Instruct-v0.2, to obtain labels for our dataset. Below we include samples of the zero-shot prompts used to obtain these results.\\

\noindent \textit{Example prompt for article-level type prediction}\\
\break
\noindent \{\lq role': \lq user', \lq content': \lq You are a helpful annotation assistant. Your task is to answer a multiple choice question based on the below information from a U.S. news article about the economy:\\
\break
\noindent So for instance the following: \\
excerpt: [article text] \\

multiple choice question: What is the main type of economic information covered in this article? \\
A. Firm-specific\\
B. Industry-specific\\
C. Macroeconomic / General Economic Conditions\\
D. Government revenue and expenses\\
E. None of the above\\

\noindent Please answer with a single letter without explanations. If you are unsure, please guess.'\}\\
\break
\textit{Example prompt for quantity-level type prediction}\\
\break
\{\lq role': \lq user',
\lq content': \lq You are a helpful annotation assistant. Your task is to answer a multiple choice question based on the below information from a U.S. news article about the economy.\\
\break
So for instance the following:\\
excerpt: [indicator text]\\
context: [context text]\\

\noindent multiple choice question: The excerpt should contain an economic indicator value. Based on the context, what type of indicator is it?\\
A. Macroeconomic / General Economic Conditions\\
B. Industry-specific\\
C. Government revenue and expenses\\
D. Personal\\
E. Firm-specific\\
F. None of the above\\
\break
\noindent Please answer with a single letter without explanations. If you are unsure, please guess.'\}

The prompts were used for each sample in the test set of each fold. For two-shot prompting, two examples were randomly selected from the training set and enveloped into the prompt as shown below.\\

\noindent \textit{Example two-shot prompt for article-level type prediction}\\
\break
\noindent \{\lq role': \lq user', \lq content': \lq You are a helpful annotation assistant. Your task is to answer a multiple choice question based on the below information from a U.S. news article about the economy:\\
\break
\noindent So for instance the following: \\
excerpt: [example 1 article text] \\

\noindent multiple choice question: What is the main type of economic information covered in this article? \\
A. Firm-specific\\
B. Industry-specific\\
C. Macroeconomic / General Economic Conditions\\
D. Government revenue and expenses\\
E. None of the above\\

\noindent Please answer with a single letter without explanations. If you are unsure, please guess.'\}\\
\break
\noindent \{\lq role': \lq assistant', \lq content': [example 1 gold label]' \}\\

\noindent \{\lq role': \lq user', \lq content': \lq You are a helpful annotation assistant. Your task is to answer a multiple choice question based on the below information from a U.S. news article about the economy:\\
\break
\noindent So for instance the following: \\
excerpt: [example 2 article text] \\

\noindent multiple choice question: What is the main type of economic information covered in this article? \\
A. Firm-specific\\
B. Industry-specific\\
C. Macroeconomic / General Economic Conditions\\
D. Government revenue and expenses\\
E. None of the above\\

\noindent Please answer with a single letter without explanations. If you are unsure, please guess.'\}\\
\break
\noindent \{\lq role': \lq assistant', \lq content': [example 2 gold label]' \}\\

\noindent \{\lq role': \lq user', \lq content': \lq You are a helpful annotation assistant. Your task is to answer a multiple choice question based on the below information from a U.S. news article about the economy:\\
\break
excerpt: [article text] \\

\noindent multiple choice question: What is the main type of economic information covered in this article? \\
A. Firm-specific\\
B. Industry-specific\\
C. Macroeconomic / General Economic Conditions\\
D. Government revenue and expenses\\
E. None of the above\\

\noindent Please answer with a single letter without explanations. If you are unsure, please guess.'\}\\

\noindent \textit{Example two-shot prompt for quantity-level type prediction}\\
\break
\{\lq role': \lq user',
\lq content': \lq You are a helpful annotation assistant. Your task is to answer a multiple choice question based on the below information from a U.S. news article about the economy.\\
\break
So for instance the following:\\
excerpt: [example 1 indicator text]\\
context: [example 1 context text]\\

\noindent multiple choice question: The excerpt should contain an economic indicator value. Based on the context, what type of indicator is it?\\
A. Macroeconomic / General Economic Conditions\\
B. Industry-specific\\
C. Government revenue and expenses\\
D. Personal\\
E. Firm-specific\\
F. None of the above\\
\break
\noindent Please answer with a single letter without explanations. If you are unsure, please guess.'\}\\

\noindent \{\lq role': \lq assistant', \lq content': [example 1 gold label]' \}\\

\{\lq role': \lq user',
\lq content': \lq You are a helpful annotation assistant. Your task is to answer a multiple choice question based on the below information from a U.S. news article about the economy.\\
\break
So for instance the following:\\
excerpt: [example 2 indicator text]\\
context: [example 2 context text]\\

\noindent multiple choice question: The excerpt should contain an economic indicator value. Based on the context, what type of indicator is it?\\
A. Macroeconomic / General Economic Conditions\\
B. Industry-specific\\
C. Government revenue and expenses\\
D. Personal\\
E. Firm-specific\\
F. None of the above\\
\break
\noindent Please answer with a single letter without explanations. If you are unsure, please guess.'\}\\

\noindent \{\lq role': \lq assistant', \lq content': [example 2 gold label]' \}\\

\{\lq role': \lq user',
\lq content': \lq You are a helpful annotation assistant. Your task is to answer a multiple choice question based on the below information from a U.S. news article about the economy.\\
\break
excerpt: [indicator text]\\
context: [context text]\\

\noindent multiple choice question: The excerpt should contain an economic indicator value. Based on the context, what type of indicator is it?\\
A. Macroeconomic / General Economic Conditions\\
B. Industry-specific\\
C. Government revenue and expenses\\
D. Personal\\
E. Firm-specific\\
F. None of the above\\
\break
\noindent Please answer with a single letter without explanations. If you are unsure, please guess.'\}\\

\paragraph{Article-level RoBERTa base classifier architecture details} For article-level classifiers, we add a classifier on top of the CLS token. For quantity-level classifiers, we use the sentence containing the excerpt and the two surrounding sentences, to the left and right. We then concatenate the CLS embedding of this contextual information with the average embedding (final-layer) of all tokens within the excerpt containing the quantity before passing it to the classifier. All base classifiers are

\paragraph{Training Settings for RoBERTa base classifiers}
The noisy data points were used to augment the training set when the corresponding "Agreed" split (see Table \ref{tab:split_counts}) was used for testing. To prevent data leakage, noisy quantitative annotations were removed from the set if their corresponding article was included in the test split. We experimented with both appending all noisy points to the training set and choosing one "best" annotation for each uniqe value (i.e. one annotation for Type was selected per article). We considered the best annotation the one contributed by the annotator with the highest agreement rate. We tested both settings using RoBERTa with and without DAPT. These results are shown in Table \ref{tab:roberta_ablation}. The best-performing trained model (bolded) was used to generate priors for the relational model. 

\begin{table}[ht]
    \centering
    \resizebox{\columnwidth}{!}{%
    \begin{tabular}{l|lll|lll}
    \toprule
     \multirow{2}{*}{Model}  & \multicolumn{3}{c}{Article-level} & \multicolumn{3}{c}{Quantity-level} \\
     & Type & Cond & Dir & Type & Ind & $+/-$ \\
     \midrule
     \textbf{Random} & 0.136 & 0.148 & 0.19 & 0.111 & 0.064 & 0.324 \\
    \textbf{Majority Label} & 0.237 & 0.269 & 0.252 & 0.193 & 0.08 & 0.382 \\
    \textbf{RoBERTa} & \textbf{0.515} & \textbf{0.697} & 0.446 & 0.642 & 0.79 & 0.796 \\
    \textbf{RoBERTa + all noisy data} & 0.411 & 0.618 & \textbf{0.493} & 0.639 & 0.793 & 0.796 \\
    \textbf{RoBERTa + best noisy data point} & 0.425 & 0.479 & 0.321 & 0.685 & 0.824 & 0.788 \\
    \textbf{RoBERTa + DAPT} & 0.382 & 0.636 & 0.47 & 0.697 & 0.769 & 0.776 \\
    \textbf{RoBERTa + DAPT + all noisy data} & 0.474 & 0.574 & 0.475 & 0.722 & 0.817 & 0.803 \\
    \textbf{RoBERTa + DAPT + best noisy data point} & 0.436 & 0.396 & 0.328 & \textbf{0.731} & \textbf{0.826} & \textbf{0.812}\\ 
     \bottomrule
    \end{tabular}}
    \caption{General Results: Average Macro F1 Scores after 5-Fold Cross Validation}
    \label{tab:roberta_ablation}
\end{table}

\subsection{Additional Results for Frame Prediction}\label{app:additional}

In Table~\ref{tab:general_weighted}, we include the weighted F1 results for our experiments. 

\begin{table}[t]
    \centering
    \resizebox{\columnwidth}{!}{%
    \begin{tabular}{l|rrr|rrr}
    \toprule
     \multirow{2}{*}{Model}  & \multicolumn{3}{c}{Article-level} & \multicolumn{3}{c}{Quantity-level} \\
     & Type & Cond & Dir & Type & Ind & $+/-$ \\
     \midrule
     \textbf{Random} & 0.266 & 0.2 & 0.243 & 0.203 & 0.107 & 0.33 \\
    \textbf{Majority Label} & 0.607 & 0.412 & 0.382 & 0.542 & 0.318 & 0.393 \\
    \textbf{Base Classifier} & 0.792 & 0.732 & 0.596 & 0.925 & 0.916 & 0.81 \\
    \textbf{Base Classifier + DAPT} & 0.785 & 0.675 & 0.563 & 0.938 & 0.913 & 0.815 \\
    \textbf{Relational (best)} & 0.778 & 0.732 & 0.596 & 0.93 & 0.922 & 0.817\\
     \bottomrule
      
    \end{tabular}}
    \caption{General Results: Average Weighted F1 Scores after 5-Fold Cross Validation}
    \label{tab:general_weighted}
\end{table}
\subsection{Addressing Concerns with Data Leakage and DAPT} We chose to include the annotated subset of data for DAPT because, while one may not have access to a labeled dataset, anyone utilizing the framework would have access to their full unlabelled dataset, and thus would be able to perform additional rounds of DAPT using self-supervised objectives. 

However, to address any concerns with data leakage, we executed DAPT with all labelled articles removed from the training set. Specifically, any articles used at any point for fine-tuning the classifiers were omitted. We present results for all versions of the model which included DAPT with this checkpoint in Tab. \ref{tab:roberta_ablation_no_leak}. We find that there is no significant advantage to including labelled articles in the pre-training set.

\begin{table}[ht]
    \centering
    \resizebox{\columnwidth}{!}{%
    \begin{tabular}{l|lll|lll}
    \toprule
     \multirow{2}{*}{Model}  & \multicolumn{3}{c}{Article-level} & \multicolumn{3}{c}{Quantity-level} \\
     & Type & Cond & Dir & Type & Ind & $+/-$ \\
     \midrule
    \textbf{RoBERTa + DAPT} & 0.382 & 0.636 & 0.47 & 0.697 & 0.769 & 0.776 \\
    \textbf{RoBERTa + DAPT (LAR)} & 0.519 & 0.521 & 0.426 & 0.700 & 0.722 & 0.763\\
    \midrule
    \textbf{RoBERTa + DAPT + all noisy data} & 0.474 & 0.574 & 0.475 & 0.722 & 0.817 & 0.803 \\
    \textbf{RoBERTa + DAPT (LAR) + all noisy data} & 0.465 & 0.562 & 0.429 & 0.698 & 0.787 & 0.776\\
    \midrule
    \textbf{RoBERTa + DAPT + best noisy data point} & 0.436 & 0.396 & 0.328 & 0.731 & 0.826 & 0.812\\ 
    \textbf{RoBERTa + DAPT (LAR) + best noisy data point} & 0.446 & 0.487 & 0.382 & 0.688 & 0.870 & 0.821\\ 
     \bottomrule
    \end{tabular}}
    \caption{Base models trained with DAPT, labelled articles removed (LAR) from pre-training set}
    \label{tab:roberta_ablation_no_leak}
\end{table}

\subsection{Error Analysis for Frame Prediction}\label{app:error_analysis}

In Tab. \ref{tab:results_by_publisher}, we observe particular instability for predictions over articles from \textit{Fox News} and the \textit{Wall Street Journal}. 

For Fox News articles, performance for article-level predictions are particularly low. We attribute this to the fact that there are only three training examples in the data. This prohibits the model from learning to make predictions for the slightly different writing style. Additionally, this means that the test set is very small and one incorrect prediction decreases the macro f1 score significantly. We note that the training example for which an incorrect article type was predicted, had a gold label of government but was predicted macro. This was also a common disagreement among annotators, as described in App.~\ref{app:cm_ann}, and therefore a more difficult problem. 

Quantity-level polarity was also unstable for Fox News. There were only two training examples for polarity. For the single error, the model predicted neutral while the gold label was negative. This was also a common disagreement among annotators. 

\begin{figure}[ht]
    \centering
    \includegraphics[width=\columnwidth]{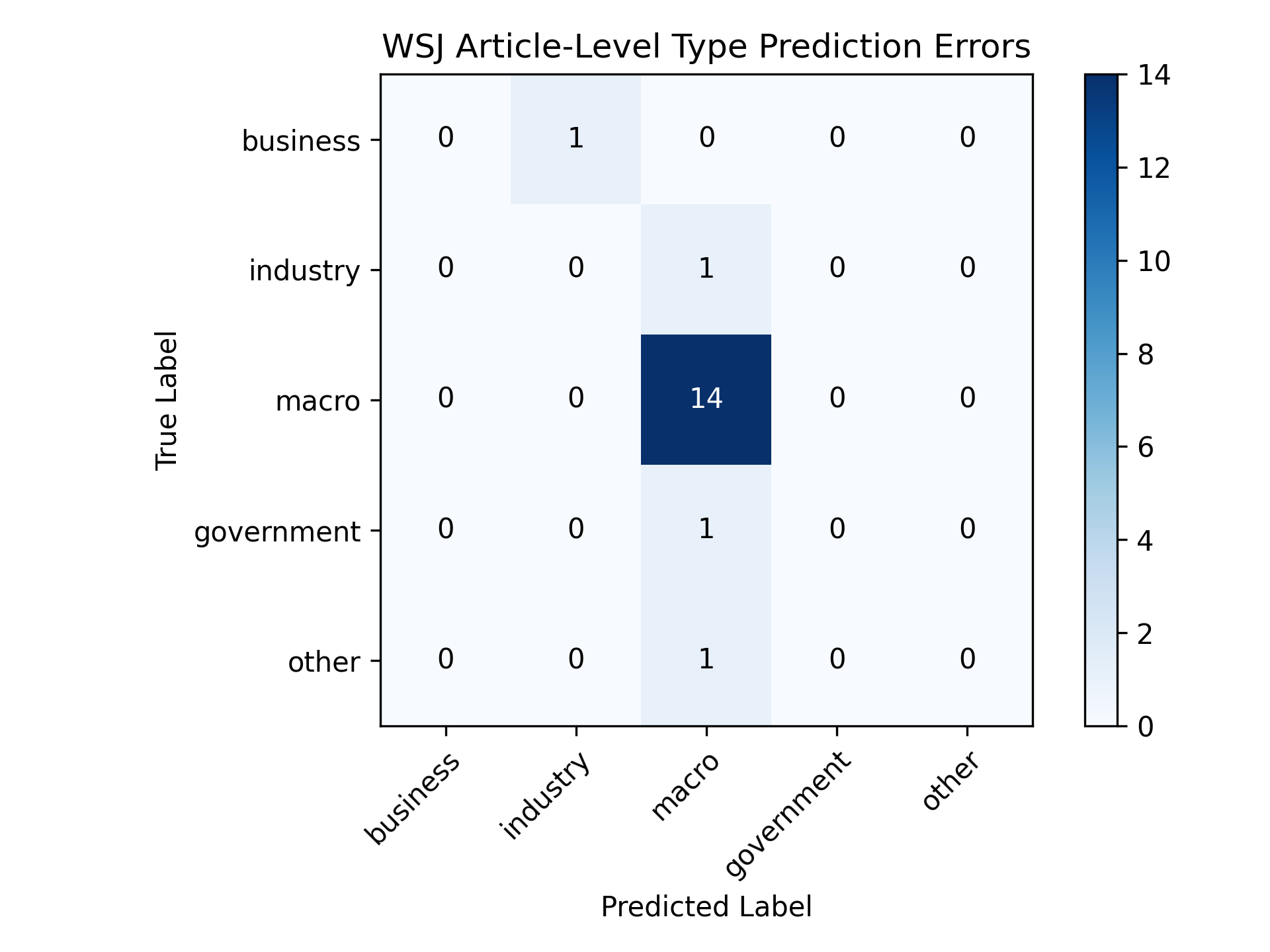}
    \caption{A significant majority of Wall Street Journal training articles were of type macro, causing the model to be biased toward this label.}
    \label{fig:apx:wsj_errors}
\end{figure}

In spite of sufficient training examples, we achieve relatively low f1 scores on article-level predictions for documents from the \textit{Wall Street Journal}. We believe that this may be attributed to the difference in format of these articles compared to those of other publishers. In Fig.~\ref{fig:apx:wsj_errors}, we observe that of the 18 training examples, 14 had macro gold labels. We believe that the class imbalance caused the model to associate the \textit{Wall Street Journal} writing with a macro label only. This is supported by the fact that the model predicted that 17 of the 18 articles were of type macro. 

Because the relational model predicts \textit{None} for the other article-level annotation components when a label other than \textit{Macro} is predicted for article type, this type of error contributes to the lower F1 scores for article-level condition and direction. We believe these errors can also be attributed to the somewhat dryer delivery of information in \textit{WSJ} articles. Empirically, their articles tend to express less opinion when discussing the economy compared to other publishers in the corpus.

\subsection{Additional Framing Analysis Graphs}\label{app:analysis_graphs}

In Section \ref{sec:analysis} we investigated how economic indicators are used by the \emph{New York Times}, the \emph{Washington Post}, and the \emph{Wall Street Journal}. In addition to the figures shown in that section we generated a number of additional graphs that can be seen below. These graphs are meant to provide additional detail on how each publication is operating. 

\begin{figure}[ht]
    \centering
    \includegraphics[width=\columnwidth]{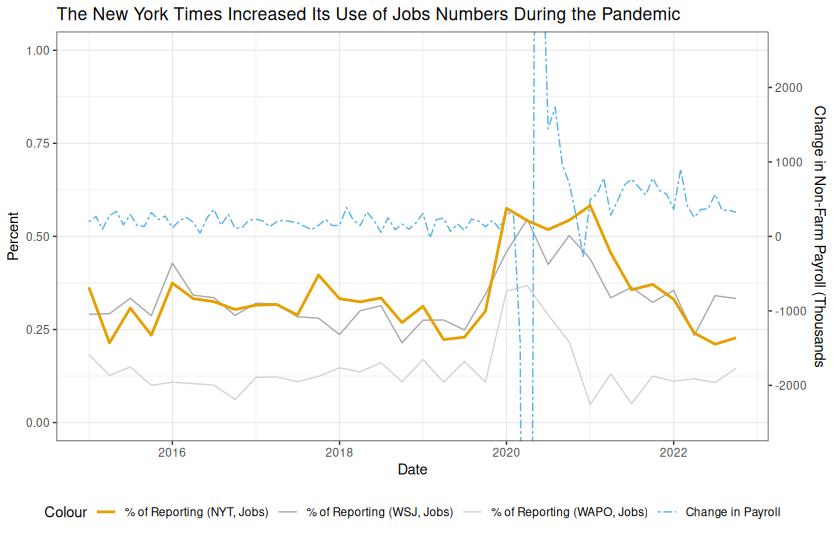}
    \caption{Selection of economic indicators referencing jobs numbers from 2015 through 2023 in the New York Times, Washington Post, and Wall Street Journal. Aggregated Quarterly. Monthly payroll data can be seen in the dotted blue line.}
    \label{fig:apx:jobs}
\end{figure}

In Figure \ref{fig:apx:jobs}, we can see how the three publications choose to use jobs indicators in relation to the underlying jobs numbers. Notably, we can see that the \emph{New York Times} and \emph{Wall Street Journal} both exhibit similar changes in their use of job indicators in response to the change in underlying data. The \emph{Washington Post} also has a notable change in behavior, but its response is less persistent than that of the other papers. 

\begin{figure}[ht]
    \centering
    \includegraphics[width=\columnwidth]{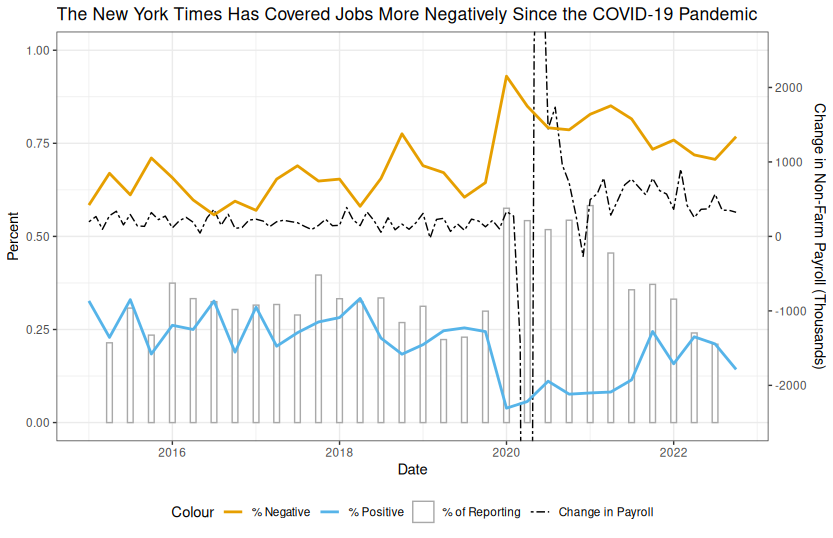}
    \caption{Framing of articles referencing job numbers from 2015 through 2023 in the New York Times. Aggregated Quarterly. Monthly payroll data can be seen in the dotted black line.}
    \label{fig:apx:nyt:jobs}
\end{figure}

\begin{figure}[ht]
    \centering
    \includegraphics[width=\columnwidth]{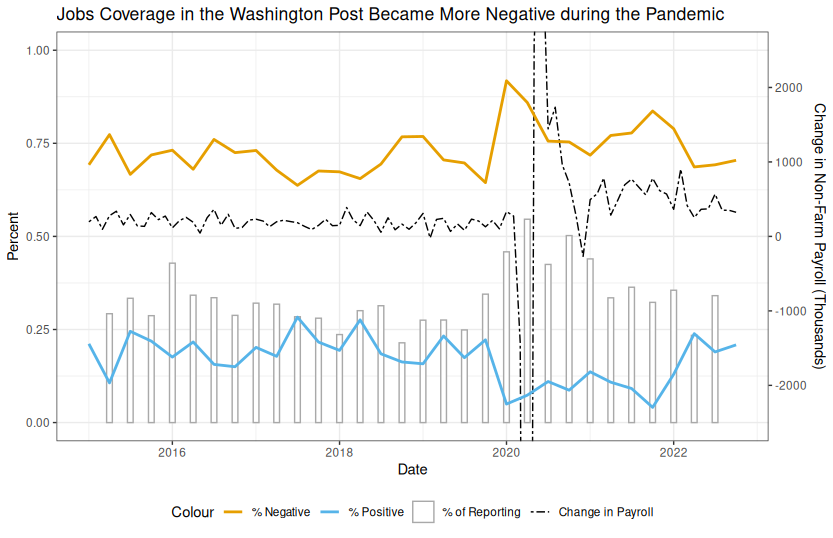}
    \caption{Framing of articles referencing job numbers from 2015 through 2023 in the Washington Post. Aggregated Quarterly. Monthly payroll data can be seen in the dotted black line.}
    \label{fig:apx:wapo:jobs}
\end{figure}

\begin{figure}[ht]
    \centering
    \includegraphics[width=\columnwidth]{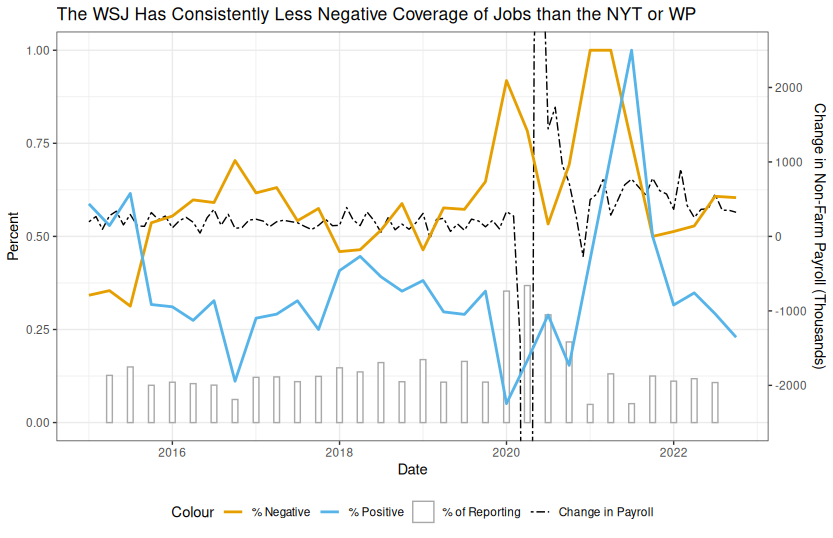}
    \caption{Framing of articles referencing job numbers from 2015 through 2023 in the Wall Street Journal. Aggregated Quarterly. Monthly payroll data can be seen in the dotted black line.}
    \label{fig:apx:wsj:jobs}
\end{figure}

Figures \ref{fig:apx:nyt:jobs}, \ref{fig:apx:wapo:jobs}, and \ref{fig:apx:wsj:jobs}, show how the \emph{New York Times}, the \emph{Washington Post} and the \emph{Wall Street Journal} use jobs indicators in their reporting.

While all three publications chose to promptly respond to the massive job loss caused by the pandemic, dramatically increasing their percent of coverage devoted to jobs, the Journal dipped back down quickly, the Post slowly dipped to their pre-pandemic norm, and the Times slowly fell below pre-pandemic norms (and as shown in Figure \ref{fig:prices}, the Times had a relatively higher surge in their coverage of prices in this post-pandemic era). Further, it is not just quantity, but the spin that makes the frame: the Journal has very malleable framing, with heavily positive coverage of jobs numbers as they soured in the early Biden administration, while the Times coverage continued to be heavily negative despite the historic run of jobs recovery.

\begin{figure}[ht!]
    \centering
    \includegraphics[width=\columnwidth]{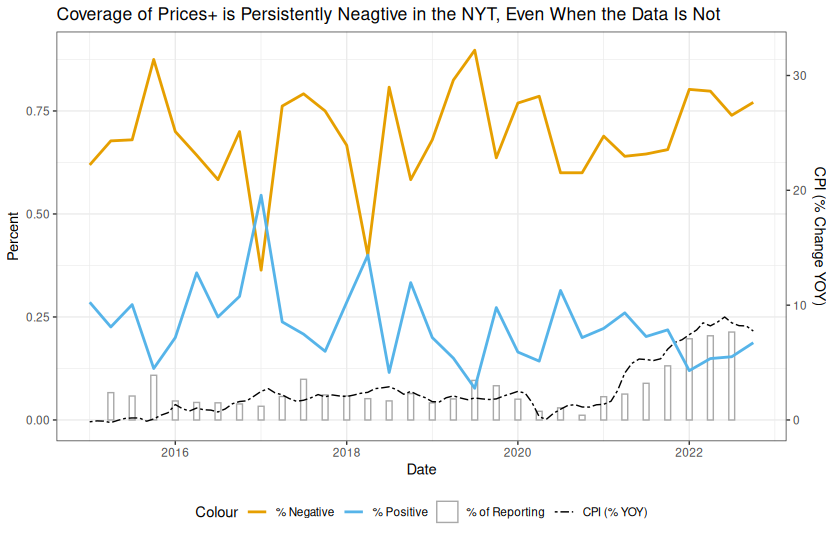}
    \caption{Framing of articles referencing price (price \& energy) numbers from 2015 through 2023 in the New York Times. Aggregated Quarterly. Monthly CPI data can be seen in the dotted black line.}
    \label{fig:apx:nyt:prices}
\end{figure}

\begin{figure}[ht!]
    \centering
    \includegraphics[width=\columnwidth]{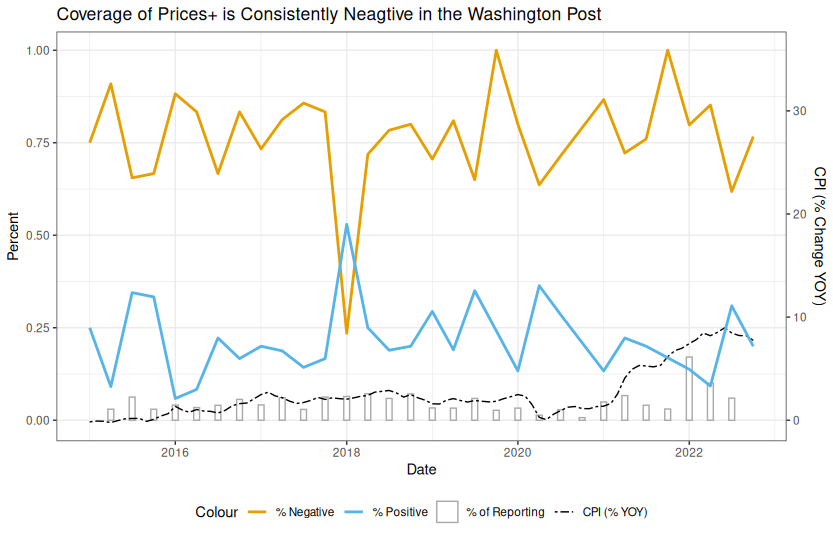}
    \caption{Framing of articles referencing price (price \& energy) numbers from 2015 through 2023 in the Washington Post Aggregated Quarterly. Monthly CPI data can be seen in the dotted black line.}
    \label{fig:apx:wapo:prices}
\end{figure}

\begin{figure}[ht!]
    \centering
    \includegraphics[width=\columnwidth]{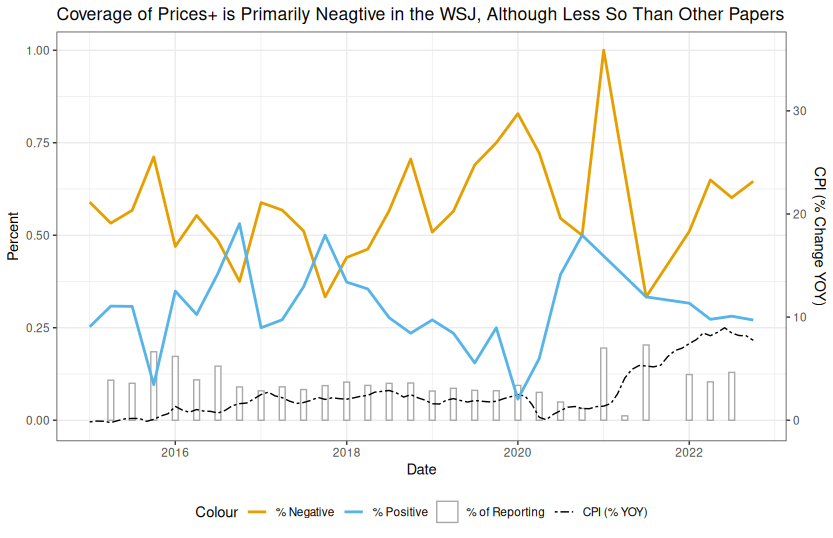}
    \caption{Framing of articles referencing price (price \& energy) numbers from 2015 through 2023 in the Wall Street Journal. Aggregated Quarterly. Monthly CPI data can be seen in the dotted black line.}
    \label{fig:apx:wsj:prices}
\end{figure}

Figures \ref{fig:apx:nyt:prices}, \ref{fig:apx:wapo:prices}, and \ref{fig:apx:wsj:prices}, show how the \emph{New York Times}, the \emph{Washington Post}, and the \emph{Wall Street Journal} use price (including energy price) indicators in their reporting. Once again, the \emph{New York Times} and \emph{Washington Post} display consistently negative framing of price indicators, while the \emph{Wall Street Journal} has a more balanced framing of such indicators. It is interesting that the framing of price indicators in the \emph{New York Times} and \emph{Washington Post} seems almost static, with no obvious change following the rise in inflation. This differs from the shifts in framing we see in both papers in Figures \ref{fig:apx:nyt:jobs} and \ref{fig:apx:wapo:jobs}.

\end{document}